\documentclass[12pt]{article}
\pdfoutput=1
\usepackage{setspace}
\usepackage[top=1in, bottom=1in, left=1in, right=1in]{geometry}
\usepackage{enumitem} 
\setlist{nolistsep} 

\usepackage{mathtools,amssymb,amsthm,amsthm,amsfonts,latexsym,bbm}
\usepackage{amscd,amsxtra} 
\usepackage{graphics,graphicx,xcolor}
\usepackage{ifpdf}
\usepackage[caption=false]{subfig}
\usepackage{multirow}
\usepackage[comma]{natbib}
\usepackage{comment}
\usepackage{ulem}
\usepackage{authblk} 
\usepackage[hypertexnames=true,colorlinks,linkcolor=blue,citecolor=blue,urlcolor=blue]{hyperref}

\theoremstyle{plain}
\newtheorem{theorem}{\sc Theorem}

\newtheorem{prop}[theorem]{\sc Proposition}

\theoremstyle{remark}

\theoremstyle{definition}
\newtheorem{definition}{\sc Definition}
\newtheorem{assumption}{\sc Assumption}

\newtheorem{algorithm}{\sc Algorithm}

\def\xv{\boldsymbol x}

\def\Sv{\boldsymbol S}

\def\Vv{\boldsymbol V}
\def\Wv{\boldsymbol W}
\def\Xv{\boldsymbol X}

\def\Zv{\boldsymbol Z}

\newcommand{\Cc}{\mathcal{C}}

\newcommand{\Wc}{\mathcal{W}}
\newcommand{\Xc}{\mathcal{X}}

\newcommand{\betav}{\mbox{\boldmath{$\beta$}}}

\newcommand{\muv}{\mbox{\boldmath{$\mu$}}}

\newcommand{\xiv}{\mbox{\boldmath{$\xi$}}}

\newcommand{\omegav}{\mbox{\boldmath{$\omega$}}}
\newcommand{\Sigmav}{\mbox{\boldmath{$\Sigma$}}}
\newcommand{\Lambdav}{\mbox{\boldmath{$\Lambda$}}}

\def\1v{\mathbf 1}
\def\0v{\mathbf 0}
\def\Id{\mathbf I} 

\newcommand{\Ind}[1]{\mathbbm{1}_{\left\{ {#1} \right\} }}

\newcommand{\R}{\mathbb R}

\newcommand{\E}{\mathbb E}
\newcommand{\sgn}{\mathop{\mathrm{sign}}}
\def\Pr{\mathrm P}

\newcommand{\var}{\mathop{\rm Var}}

\newcommand{\ds}{\displaystyle}
\newcommand{\mb}{\mbox}
\newcommand{\wh}{\widehat}
\newcommand{\argmin}{\operatornamewithlimits{argmin}}
\newcommand{\argmax}{\operatornamewithlimits{argmax}}

\newcommand{\set}[1]{\left\{#1\right\}}

\def\half{\frac{1}{2}}

\def\ie{\textit{i.e.}}

\def\ie{\textit{i.e.}}

\title{Flexible High-dimensional Classification Machines and Their Asymptotic Properties}
\author{Xingye Qiao\thanks{Corresponding author}}
\affil{Department of Mathematical Sciences\authorcr  State University of New York, Binghamton, NY 13902-6000.\authorcr E-mail: \texttt{qiao@math.binghamton.edu}}
\author{Lingsong Zhang}
\affil{Department of Statistics\authorcr Purdue University, West Lafayette, IN 47907.\authorcr E-mail: \texttt{lingsong@purdue.edu}}

\date{}
\begin{document}
\setlength{\belowdisplayskip}{2pt} \setlength{\belowdisplayshortskip}{2pt}
\setlength{\abovedisplayskip}{2pt} \setlength{\abovedisplayshortskip}{2pt}
\maketitle
\pagenumbering{Roman}
\newpage

\begin{abstract}
Classification is an important topic in statistics and machine learning with great potential in many real applications. In this paper, we investigate two popular large margin classification methods, Support Vector Machine (SVM) and Distance Weighted Discrimination (DWD), under two contexts: the high-dimensional, low-sample size data and the imbalanced data. A unified family of classification machines, the \textsc{FL}exible \textsc{A}ssortment \textsc{M}achin\textsc{E} (\textsc{FLAME}) is proposed, within which DWD and SVM are special cases. The FLAME family helps to identify the similarities and differences between SVM and DWD. It is well known that many classifiers overfit the data in the high-dimensional setting; and others are sensitive to the imbalanced data, that is, the class with a larger sample size overly influences the classifier and pushes the decision boundary towards the minority class. SVM is resistant to the imbalanced data issue, but it overfits high-dimensional data sets by showing the undesired data-piling phenomena. The DWD method was proposed to improve SVM in the high-dimensional setting, but its decision boundary is sensitive to the imbalanced ratio of sample sizes. Our FLAME family helps to understand an intrinsic connection between SVM and DWD, and improves both methods by providing a better trade-off between sensitivity to the imbalanced data and overfitting the high-dimensional data. Several asymptotic properties of the FLAME classifiers are studied. Simulations and real data applications are investigated to illustrate the usefulness of the FLAME classifiers.

\vspace{0.5in}

\end{abstract}

\newpage
\pagenumbering{arabic}
\setcounter{page}{1}

\section{Introduction}\label{sec:intro}
Classification refers to predicting the class label, $y\in \Cc$, of a data object based on its covariates, $\xv \in \mathcal{X}$. Here $\Cc$ is the space of class labels, and $\mathcal{X}$ is the space of the covariates. Usually we consider $\mathcal{X}\equiv\mathbb{R}^d$, where $d$ is the number of variables or the dimension. See \cite {Duda2001Pattern} and \citet{hastie2009elements} for comprehensive introductions to many popular classification methods. When $\Cc=\{+1, -1\}$, this is an important class of classification problems, called binary classification. The classification rule for a binary classifier usually has the form $\phi(\xv)=\sgn\left\{f(\xv)\right\}$, where $f(\xv)$ is called the discriminant function. Linear classifiers are the most important and the most commonly used classifiers, as they are often easy to interpret in addition to reasonable classification performance. We focus on linear classifier in this article. In the above formula, linear classifiers correspond to $f(\xv;\omegav,\beta)=\xv^T\omegav+\beta$. The sample space is divided into halves by the \textit{separating hyperplane}, also known as the \textit{classification boundary}, defined by $\left\{\xv:f(\xv)\equiv\xv^T\omegav+\beta=0\right\}$. Note that the coefficient vector $\omegav\in \R^d$ defines the normal vector, and hence the direction, of the classification boundary, and the intercept term $\beta \in \R$ defines the location of the classification boundary.

In this paper, two popular classification methods, Support Vector Machine \citep[SVM;][]{Cortes1995Support, vapnik1998statistical, cristianini2000introduction} and Distance Weighted Discrimination \citep[DWD;][]{marron2007distance, Qiao2010Weighteda} are investigated under two important contexts: the High-Dimensional, Low-Sample Size (HDLSS) data and the imbalanced data. Both methods are large margin classifiers \citep{smola2000advances}, that seek separating hyperplanes which maximize certain notions of \textit{gap} (\ie, distances) between the two classes. The investigation of the performance of SVM and DWD motivates the invention of a novel family of classifiers, the \textsc{FL}exible \textsc{A}ssortment \textsc{M}achin\textsc{E} (\textsc{FLAME}), which unifies the two classifiers, and helps to understand their connections and differences.

\subsection{Motivation: Pros and Cons of SVM and DWD}
SVM is a very popular classifier in statistics and machine learning. It has been shown to have Fisher consistency, \ie, when sample size goes to infinity, its decision rule converges to the Bayes rule \citep{Lin2004note}. SVM has several nice properties. 1) Its dual formulation is relatively easy to implement (by the Quadratic Programming). 2) SVM is robust to the model specification, which makes it very popular in various real applications. However, when being applied to HDLSS data, it has been observed that a large portion of the data (usually the support vectors, to be properly defined later) lie on two hyperplanes parallel to the SVM classification boundary. This is known as the \textit{data-piling} phenomena \citep{marron2007distance,ahn2010maximal}. Data-piling of SVM indicates a type of overfitting. Other overfitting phenomena of SVM under the HDLSS context include:
\begin{enumerate}
		\item The angle between the SVM direction and the Bayes rule direction is usually large.
		\item The variability of the sampling distribution of the SVM direction $\omegav$ is very large \citep{zhang2011some}. Moreover, because the separating hyperplane is decided only by the support vectors, the SVM direction tends to be unstable, in the sense that small turbulence or measurement error to the support vectors can lead to big change of the direction.
		\item In some cases, the out-of-sample classification performance may not be optimal due to the suboptimal direction of the estimated SVM discrimination direction.
\end{enumerate}

DWD is a recently developed classifier to improve SVM in the HDLSS setting. It uses a different notion of gap from SVM. While SVM is to maximize the smallest distance between classes, DWD is to maximize a special average distance (harmonic mean) between classes. It has been shown in many earlier simulations that DWD largely overcomes the overfitting (data-piling) issue and it usually gives a better discrimination direction.

On the other hand, the intercept term $\beta$ of the DWD method is sensitive to the sample size ratio between the two classes, \ie, to the imbalanced data \citep{Qiao2010Weighteda}. Note that, even though a good discriminant direction $\omegav$ is more important in revealing the profiling difference between the two populations, the classification/prediction performance heavily depends on the intercept $\beta$, more than on the direction $\omegav$. As shown in \citet{Qiao2010Weighteda}, usually the $\beta$ of the SVM classifier is not sensitive to the sample size ratio, while the $\beta$ of the DWD method will become too large (or too small) if the sample size of the positive class (or negative class) is very large.

In summary, both methods have pros and cons. SVM has larger stochastic variability and usually overfits the data by showing data-piling phenomena, but is less sensitive to the imbalanced data issue. DWD usually overcomes the overfitting/data-piling issue, and has smaller sampling variability, but is very sensitive to the imbalanced data. Driven by their similarity, we propose a unified class of classifiers, FLAME, in which the above two classifiers are special cases. FLAME provides a framework to study the connections and differences between SVM and DWD. Each FLAME classifier has a parameter $\theta$ which is used to control the performance balance between overfitting the HDLSS data and the sensitivity to the imbalanced data. It turns out that the DWD method is FLAME with $\theta=0$; and that the SVM method corresponds to FLAME with $\theta=1$. The optimal $\theta$ depends on the trade-off among several factors: stochastic variability, overfitting and resistance against the imbalanced data. In this paper, we also propose two approaches to select $\theta$, where the resulting FLAME have a balanced performance between the SVM and DWD methods.

\subsection{Outlines}
The rest of the paper is organized as follows. Section \ref{SVMnDWD} provides toy examples and highlights the strengths and drawbacks of SVM and DWD on classifying the HDLSS and imbalanced data.  We develop the FLAME method in Section \ref{sec:flame family}, which is motivated by the investigation of the loss functions of SVM and DWD.
Section \ref{sec:parameter} provides suggestions for the parameters. Three types of asymptotic results for the FLAME classifier are studied in Section \ref{sec:theory}. Sections \ref{sec:simulations} discusses its properties using simulation experiments. A real application is discussed in Section \ref{sec:real_data}. Some concluding remarks and discussions are made in Section \ref{sec:conclude}.

\section{Comparison of SVM and DWD \label{SVMnDWD}}
In this section, we use several toy examples to illustrate the strengths and drawbacks of SVM and DWD under two contexts: HDLSS data and imbalanced data.

\subsection{Overfitting HDLSS data}\label{overfitting}
We use simulations to compare SVM and DWD. The results show that the stochastic variability of the SVM direction is usually larger than that of the DWD method, and SVM directions are deviated farther away from Bayes rule directions. In addition, the new proposed FLAME machine (see details in Section \ref{sec:flame family}) is also included in the comparison, and it turns out that FLAME is between the other two.
\begin{figure}[!ht]
	\begin{center}
		\includegraphics[height=0.8\linewidth, width=0.56\linewidth,angle=270]{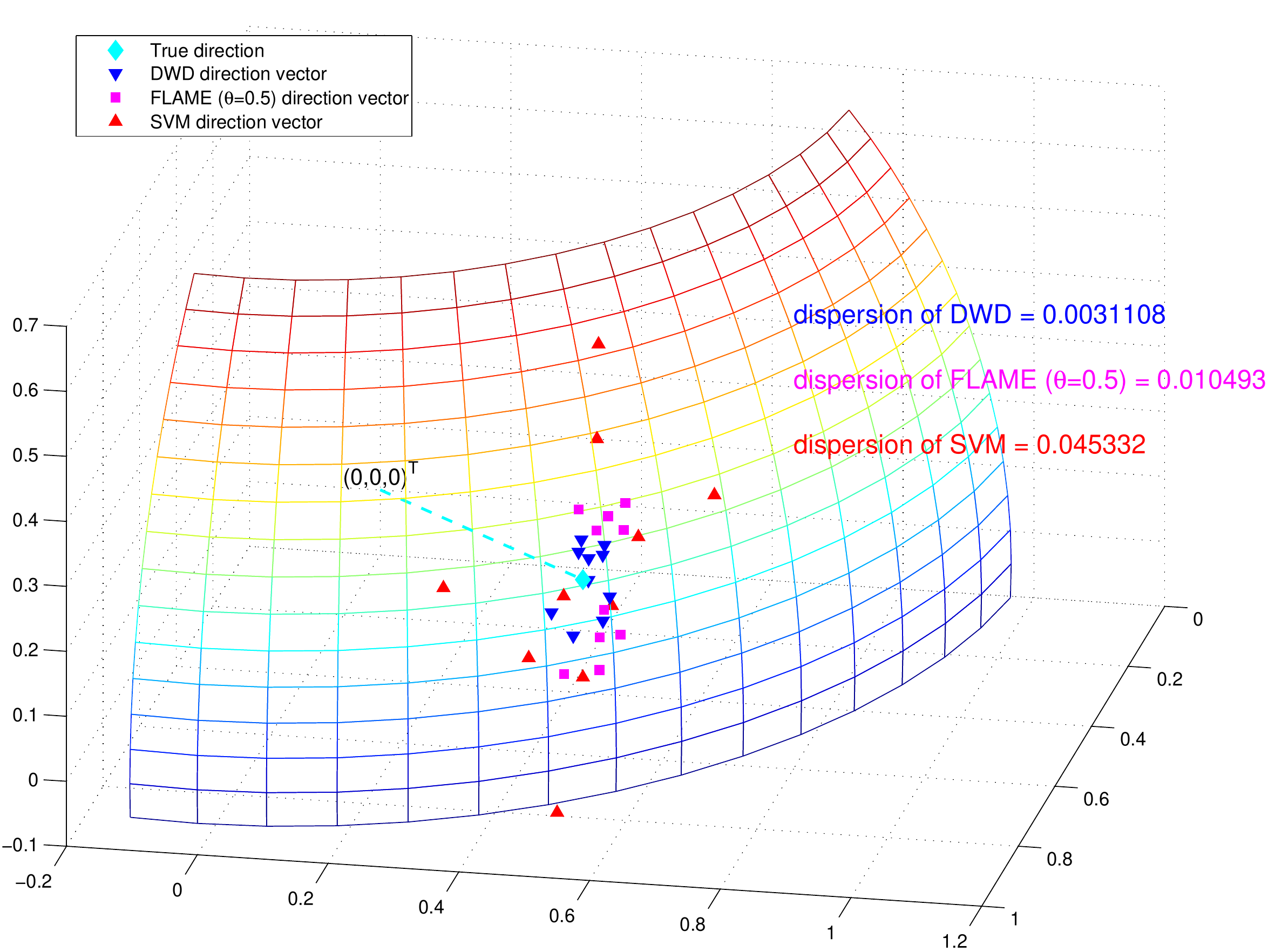}
	\end{center}
	\caption{The true population mean difference direction vector (the cyan dashed line and diamond marker; equivalent to the Bayes rule direction), the DWD directions (blue down-pointing triangles), the FLAME directions with $\theta=0.5$ (magenta squares), and the SVM directions (red up-pointing triangles) for 10 realizations of simulated data. Each direction vector has norm 1 and thus is depicted as a point on the 3D unit sphere. On average, all machines have their discriminant direction vectors scattering around the true direction. The DWD directions are the closest to the true direction and have a smaller variation. The SVM directions have the largest variation and are farthest from the true direction. The variation of the intermediate FLAME direction vectors is between the two machines above. The variation of a machine is also measured by the trace of the sample covariance calculated from the 10 resulting direction vectors for the 10 simulations.}
	\label{fig:unit_sphere}
\end{figure}

Figure \ref{fig:unit_sphere} shows the comparison results between SVM, DWD and FLAME (with some chosen tuning parameters). We simulate 10 samples with the same underlying distribution. Each simulated data set contains 12 variables and two classes, with 120 observations in each class. The two classes have mean difference on only the first three dimensions and the within-class covariances are diagonal, that is, the variables are independent. For each simulated data set, we plot the first three components of the resulting discriminant directions from SVM, DWD and FLAME (after normalizing the 3D vectors to have unit norms), as shown in Figure \ref{fig:unit_sphere}. It clearly shows that the DWD directions (the blue down-pointing triangles) are the closest ones to the \textit{true} Bayes rule direction (shown as the cyan diamond marker) among the three approaches. In addition, the DWD directions have a smaller variation (\ie, more stable) over different samples. The SVM directions (the red up-pointing triangles) are farthest from the \textit{true} Bayes rule direction and have a larger variation than the other two methods. To highlight the direction variabilities of the three methods, we introduce a novel measure for the variation (unstableness) of the discriminant directions: the trace of the sample covariance of the resulting direction vectors over the 10 replications, which we name as \textit{dispersion}. The dispersion for the DWD method (0.0031) is much smaller than that of the SVM method (0.0453), as highlighted in the figure as well. The new FLAME classifiers usually have the performance between DWD and SVM. Figure \ref{fig:unit_sphere} shows the results of a specific FLAME ($\theta=0.5$, the magenta squares), which is better than SVM but worse than DWD.

\begin{figure}[!ht]
  \begin{center}
   \includegraphics[height=0.72\linewidth, width=0.48\linewidth, angle=270]{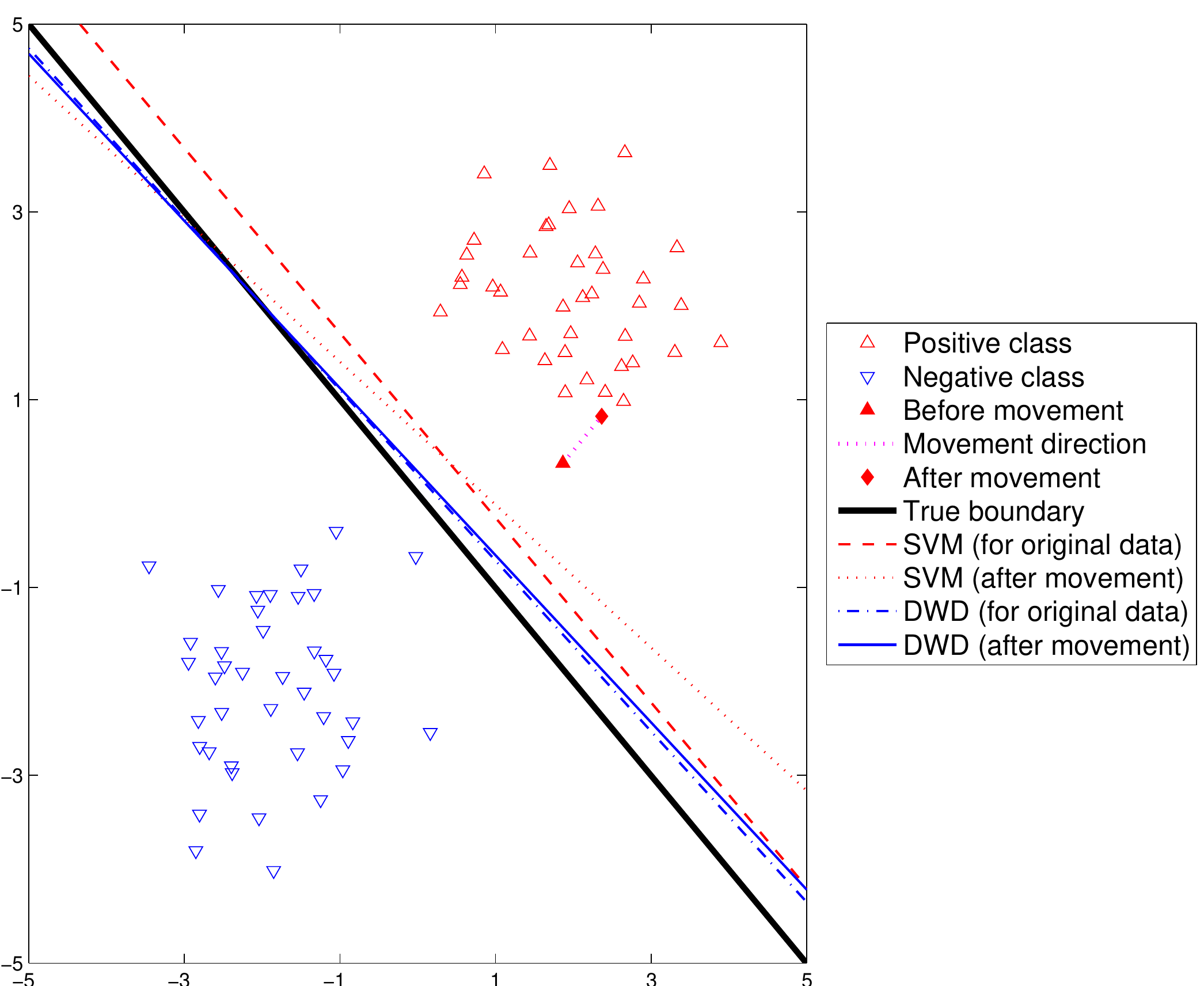}
  \end{center}
  \caption{A 2D example shows that the unstable SVM boundary has changed due to a small turbulence of a support vector (the solid red triangle and diamond) while the DWD boundary remains almost still.}
 \label{fig:example_stable}
\end{figure}

Besides the stochastic variability and the deviation from the true direction comparisons shown above, DWD outperforms SVM in terms of stability in the presence of small perturbations applied to some observations. In Figure \ref{fig:example_stable}, we use a two-dimensional example to illustrate this phenomenon. We simulate a perfectly separable 2-dimensional data set. The theoretical Bayes rule decision boundary is shown as the thick black line. The dashed red line  and the dashed dotted blue line are the SVM and the DWD classification boundaries before the perturbation. We then move one observation in the positive group a little (from the solid triangle to the solid diamond as shown in the figure). This perturbation leads to a large change of direction in SVM (shown as the dotted red line), but a small change for DWD (shown as the solid blue line). Note that all four hyperplanes are capable of classifying this training data set perfectly. But it may not be true for an out-of-sample test set. This example shows small perturbation may lead to unstableness in SVM.

\subsection{Sensitivity to imbalanced data}
In the last subsection, we have shown that DWD outperforms SVM in estimating the discrimination direction, that is, DWD directions are closer to the Bayes rule discrimination directions and have smaller variability. However, it was found that the location of DWD classification boundary, which is characterized by the intercept $\beta$, is sensitive to the sample size ratio between the two classes \citep{Qiao2010Weighteda}.

Usually, a good discriminant direction $\omegav$ helps to reveal the profiling difference between two classes of populations. But the classification/prediction performance heavily depends on the location coefficient $\beta$. We define the \textit{imbalance factor} $m\geq 1$ as the sample size ratio between the majority class and the minority class. It turns out that $\beta$ in the SVM classifier is not sensitive to $m$. However, the $\beta$ for the DWD method is very sensitive to $m$. We also notice that, as a consequence, the DWD separating hyperplane will be pushed toward the minority class, when the  ratio $m$ is close to infinity, \ie, DWD classifiers intend to ignore the minority class.  Again, we use a toy example in order to better illustrate the impact of the imbalanced data on $\beta$ and on the classification performance.

\begin{figure}[!t]
  \begin{center}
   \includegraphics[height=0.8\linewidth, width=0.56\linewidth, angle=270]{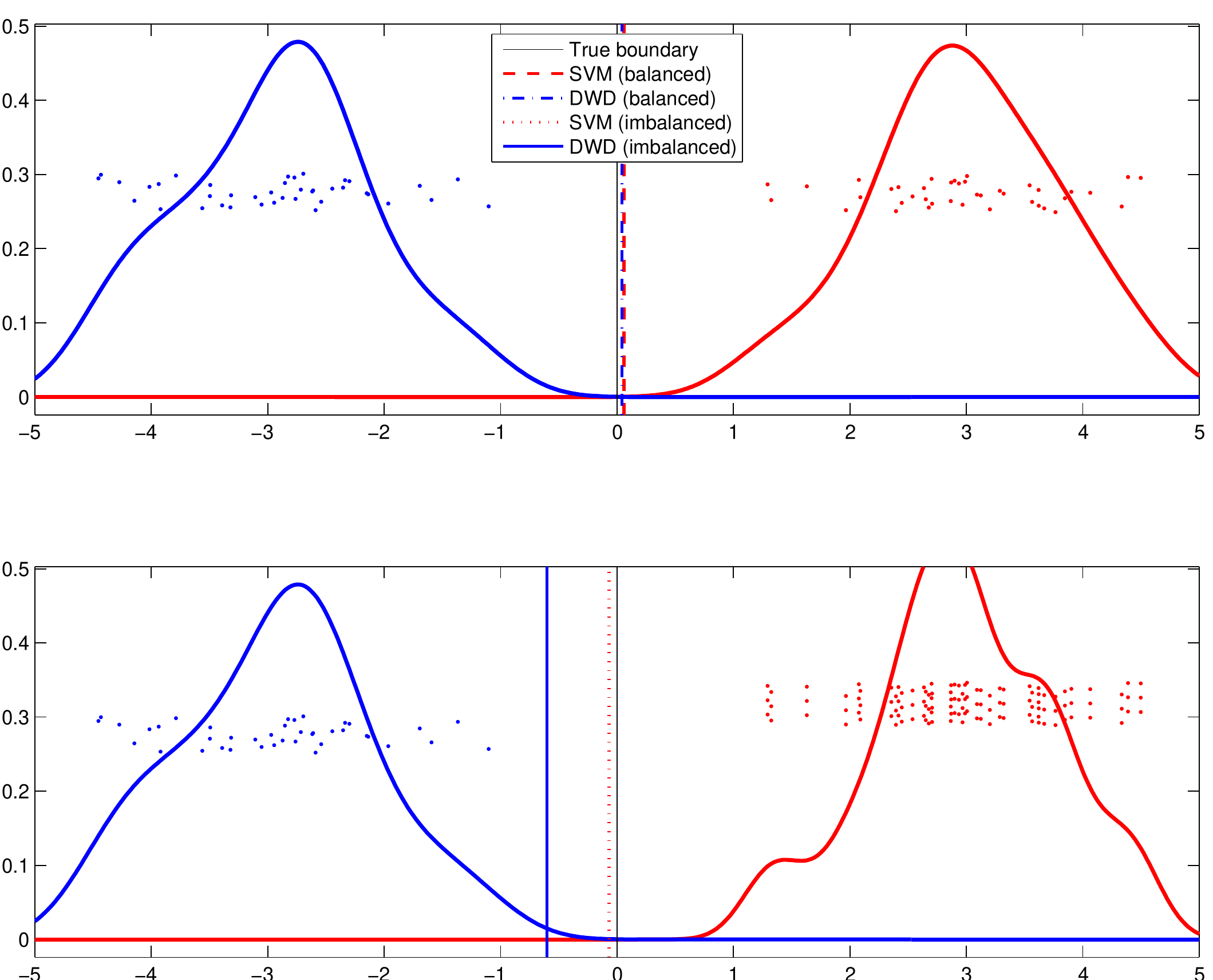}
  \end{center}
  \caption{A 1D example shows that the DWD boundary is pushed towards the minority class (blue) when the majority class (red) has tripled its sample size.}
 \label{fig:example_imbalance}
\end{figure}

Figure \ref{fig:example_imbalance} uses a one-dimensional example, so that estimating $\omegav$ is not needed. This also corresponds to a multivariate data set, where $\omegav$ is estimated correctly first, after which the data set is projected to $\omegav$ to form the one-dimensional data. In this plot, the $x$-coordinates of the red dots and the blue dots  are the values of the data while the $y$-coordinates are random jitters for better visualization. The red and blue curves are the kernel density estimations for both classes. In the top subplot of Figure \ref{fig:example_imbalance}, where $m=1$ (\ie, the balanced data), both the DWD (blue lines) and SVM (red lines) boundaries are close to the Bayes rule boundary (black solid line), which sits at 0. In the bottom subplot, the sample size of the red class is tripled, which corresponds to $m=3$. Note that the SVM boundary moves a little towards the minority (blue) class, but still fairly close to the true boundary. The DWD boundary, however, is pushed towards the minority. Although this does not impose immediate problems for the training data set, the DWD classifier will suffer from a great loss of classification performance when it is applied to an out-of-sample data set. It can be shown that when $m$ goes to infinity, the DWD classification boundary will tends to negative infinity, which totally ignores the minority group (see our Theorem \ref{thm3:oDWD_intercept_diverge}). However, SVM will not suffer from severe imbalanced data problems. One reason is that SVM only needs a small fraction of data (called support vectors) for estimating both $\omegav$ and $\beta$, which mitigate the imbalanced data issue naturally.

Imbalanced data issues have been investigated in both statistics and machine learning. See an extensive survey in \citet{Chawla2004Editorial:}. Recently, \citet{owen2007infinitely} studied the asymptotic behavior of infinitely imbalanced binary logistic regression. In addition, \citet{qiao2009adaptive} and \citet{Qiao2010Weighteda}  proposed to use adaptive weighting approaches to overcome the imbalanced data issue.

In summary, the performance of DWD and SVM is different in the following ways: 1) The SVM direction usually has a larger variation and deviates farther from the Bayes rule direction than the DWD direction does, which are indicators of overfitting HDLSS data. 2) The SVM intercept is not sensitive to the imbalanced data, but the DWD intercept is. This motivates us to investigate their similarity and differences. In the next section, a new family of classifier will be proposed, which unifies the above two classifiers.

\section{FLAME Family}\label{sec:flame family}
In this section, we introduce FLAME, a family of classifiers which is motivated by a thorough investigation of the loss functions of SVM and DWD in Section \ref{sec:loss_functions}. The formulation and implementation of the FLAME classifiers are given in Section \ref{sec:FLAME}.

\subsection{SVM and DWD loss functions}\label{sec:loss_functions}
\begin{figure}[!ht]
  \begin{center}
   \includegraphics[height=0.4\linewidth, width=0.45\linewidth]{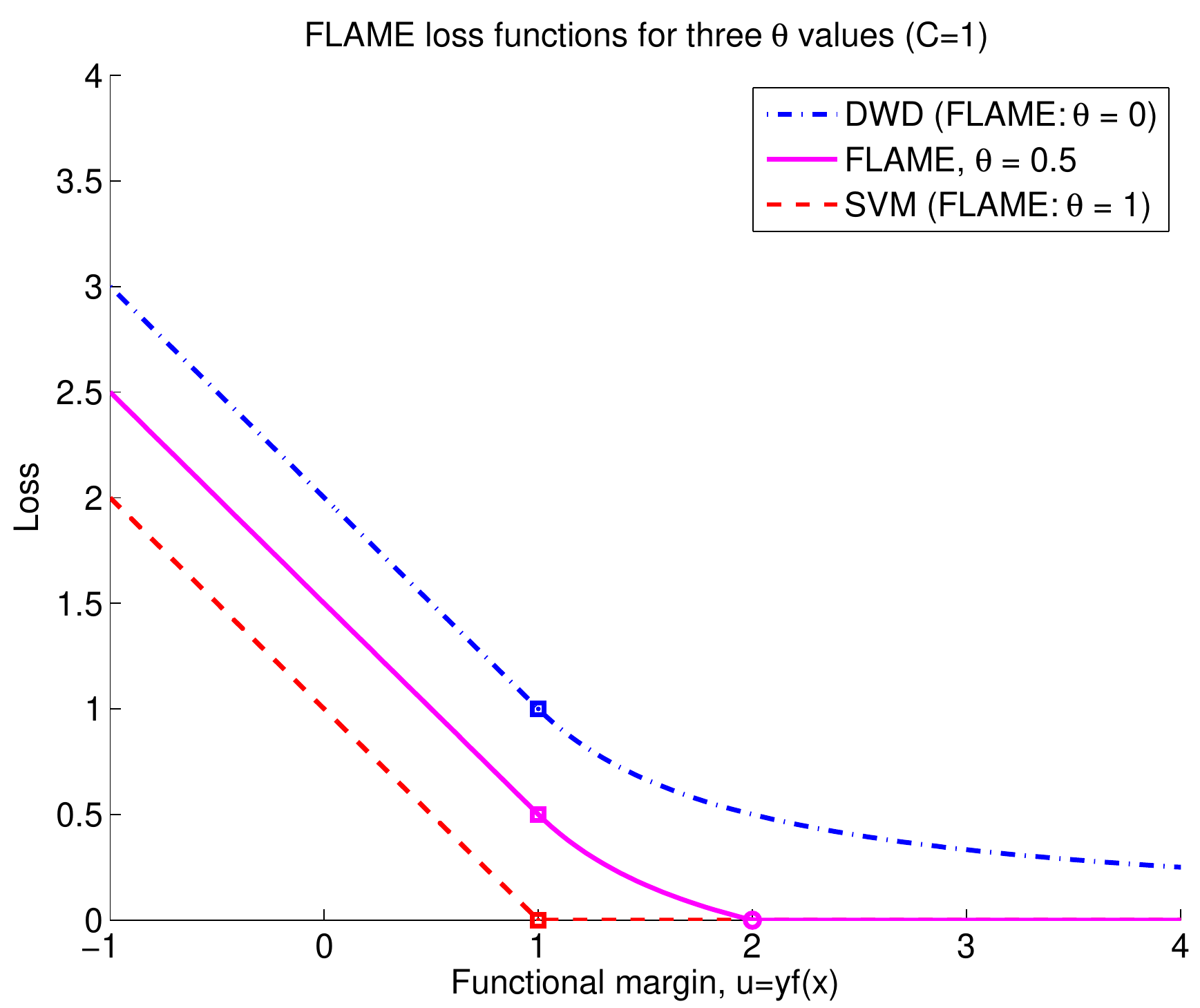}
  \end{center}
  \caption{FLAME loss functions for three $\theta$ values: $\theta=0$ (equivalent to SVM/Hinge loss), $\theta=0.5$, $\theta=1$ (equivalent to DWD). The parameter $C$ is set to be 1.}
 \label{fig:loss}
\end{figure}

The key factors that drive the very distinct performances of the SVM and the DWD methods are their associated loss functions (see Figure \ref{fig:loss}.)

Figure \ref{fig:loss} displays the loss functions of SVM, DWD and FLAME with some specific tuning parameters. SVM uses the Hinge loss function, $H(u)=(1-u)_+$ (the red dashed curve in Figure \ref{fig:loss}), where $u$ corresponds to the functional margin $u\equiv yf(\xv)$. Note that the functional margin $u$ can be viewed as the distance of vector $\xv$ from the separating hyperplane (defined by $\set{\xv:f(\xv)=0}$). When $u>0$ and is large, the data vector is correctly classified and is far from the separating hyperplane; and $u<0$, the data vector is wrongly classified. Note that when $u>1$, the corresponding Hinge loss equals zero. Thus, only those observations with $u\leq 1$ will contribute to estimating $\omegav$ and $\beta$. These observations are called \textit{support vectors}. This is why SVM is insensitive to observations that are far away from the decision boundary, and why it is less sensitive to the imbalanced data issue. However, the only influence by the support vectors makes the SVM solution subject to the overfitting (data-piling) issue. This can be explained by that the optimization of SVM would try to push vectors towards small loss, \ie, large functional margin $u$. But once a vector is pushed to the point where $u=1$, the optimization lacks further incentive to continue pushing it towards a larger function margin as the Hinge loss cannot be reduced for this vector. Therefore many data vectors are piling along the hyperplanes corresponding to $u=1$. Data-piling is bad for generalization because small turbulence to the support vectors could lead to big difference of the discriminant direction vector (recall the examples in Section \ref{overfitting}).

The DWD method corresponds to a different DWD loss function,
\begin{align}\label{dwdloss}
 V(u)=\left\{
	     \begin{array}{cc}
              2\sqrt{C}-Cu & \mb{if } u\leq \frac{1}{\sqrt{C}},\\
	      			1/u& \mb{otherwise}.
       \end{array}\right.
\end{align}
Here $C$ is a pre-defined constant.  Figure \ref{fig:loss} shows the DWD loss function with $C=1$. It is clear that the DWD loss function is very similar to the SVM loss function when $u$ is small (both are linearly decreasing with respect to $u$). The major difference is that the DWD loss is always positive. This property will make the DWD method behave in a very different way than SVM. As there is always incentive to make the function margin to be larger (and the loss to be smaller), the DWD loss function kills data-piling, and mitigates the overfitting issue for HDLSS data.

On the other hand, the DWD loss function makes the DWD method very sensitive to the imbalanced data issue, since each observation will have some influence, and thus the larger class will have larger influence. The decision boundary of the DWD method will intend to ignore the smaller class, because sacrificing the smaller class (boundary being closer to the smaller class and farther from the larger class) can lead to a dramatic reduction of the loss, which ultimately lead to a minimized overall loss.

\subsection{FLAME}\label{sec:FLAME}
We propose to borrow strengths from both methods to simultaneously deal with both the
imbalanced data and the overfitting (data-piling) issues. We first highlight the connections between the DWD loss and an modified version of the Hinge loss (of SVM). Then we modify the DWD loss so that samples
far from the classification boundary will have zero loss.

Let $f(\xv)=\xv^T\omegav+\beta$. The formulation of SVM can be rewritten (see details in the appendix) in the form of ${\ds\argmin_{\omegav,\beta}}~\sum_i H^*(y_if(\xv_i))$, $\mb{s.t. }\|\omegav\|^2\leq 1$ where the modified Hinge loss function $H^*$ is defined as
\begin{align}\label{h2loss}
 H^*(u)=\left\{
	     \begin{array}{cc}
              \sqrt{C}-Cu & \mb{if } u\leq \frac{1}{\sqrt{C}},\\
	      0 & \mb{otherwise}.
             \end{array}\right.
\end{align}
Comparing the DWD loss (\ref{dwdloss}) and this modified Hinge loss (\ref{h2loss}), one can easily see their connections: for $u\leq \frac{1}{\sqrt{C}}$, the DWD loss is greater than the Hinge loss of SVM by an exact constant $\sqrt{C}$, and for $u>\frac{1}{\sqrt{C}}$, the DWD loss is $1/u$ while the SVM Hinge loss equals 0. Clearly the modified Hinge loss (\ref{h2loss}) is the result of soft-thresholding the DWD loss at $\sqrt{C}$. In other words, SVM can be seen as a special case of DWD where the losses of those vectors with $u=y_if(\xv_i)>1/\sqrt{C}$ are shrunken to zero. To allow different levels of soft-thresholding, we propose to use a new loss function which (soft-)thresholds the DWD loss function by constant $\theta\sqrt{C}$ where $0\leq \theta\leq 1$, that is, a fraction of $\sqrt{C}$. The new loss function is
\begin{align}
	L(u)&=\left[V(u)-\theta\sqrt{C}\right]_+=\left\{
		     \begin{array}{cc}
	              (2-\theta)\sqrt{C}-Cu & \mb{if } u\leq \frac{1}{\sqrt{C}},\\
		      1/u-\theta\sqrt{C} & \mb{if }  \frac{1}{\sqrt{C}}\leq u< \frac{1}{\theta\sqrt{C}},\\
		      0& \mb{if }  u\geq   \frac{1}{\theta\sqrt{C}},\label{turning}
	             \end{array}\right.
\end{align}
that is, to reduce the DWD loss by a constant, and truncate it at 0. The magenta solid curve in Figure \ref{fig:loss} is the FLAME loss when $C=1$ and $\theta=0.5$. This simple but useful modification unifies the DWD and SVM methods. When $\theta = 1$, the new loss function (when $C=1$) reduces to the SVM Hinge loss function; while when $\theta=0$, it remains as the DWD loss.

Note that $L(u) = 0$ for $u>1/(\theta\sqrt{C})$. Thus, those data vectors with large functional margins will have zero loss. For DWD loss, because it corresponds to $\theta=0\Rightarrow 1/(\theta\sqrt{C})=\infty$, no data vector can have zero loss. For SVM loss, all the data vector with $u>1/(\theta\sqrt{C})=1/\sqrt{C}$ will have zero loss.  Training a FLAME classifier with $0<\theta<1$ can be interpreted as sampling a portion of data which are farther from the boundary than  ${1}/{\theta\sqrt{C}}$ and assign zero loss to them. Alternatively, it can be viewed as sampling data that are closer to the boundary than  ${1}/{\theta\sqrt{C}}$ and assign positive loss to them. Note that the larger $\theta$ is, the fewer data are sampled to have positive loss. As one can flexibly choose $\theta$, the new classification method with this new loss function is called the \textsc{FL}exible \textsc{A}ssortment \textsc{M}achin\textsc{E} (\textsc{FLAME}).

FLAME can be implemented by a Second-Order Cone Programming algorithm \citep{Toh1999SDPT,Tutuncu2003Solving}. Let $\theta\in [0,1]$ be the FLAME parameter. The proposed method minimizes $\ds{\min_{\omegav,b,\xiv}  \sum_{i=1}^n \left(\frac{1}{r_i}+C\xi_i-\theta\sqrt{C}\right)_+}$. A slack variable $\varphi_i\geq 0$ can be introduced to absorb the $(\cdot)_+$ function. The optimization of the FLAME can be written as
\begin{align*}
&\min_{\omegav,b,\xiv}\sum_i \varphi_i,\\
\mb{s.t.}&\quad\Big(\frac{1}{r_i}+C\xi_i-\theta\sqrt{C}\Big)-\varphi_i\leq 0,~\varphi_i\geq
0,\\
\quad&r_i=y_i(\xv_i^T\omegav+\beta)+\xi_i,~r_i\geq 0\mb{ and }\xi_i\geq 0,\\
 &\|\omegav\|^2\leq 1.
\end{align*}
A \texttt{Matlab} routine has been implemented and is available at the authors' personal websites. See the online supplementary materials for more details on the implementation.

\section{Choice of Parameters}\label{sec:parameter}
There are two tuning parameters in the FLAME model: one is the $C$, inherited from the DWD loss, which controls the amount of allowance for misclassification; the other is the FLAME parameter $\theta$, which controls the level of soft-thresholding. Similar to the discussion in DWD \citep{marron2007distance}, the classification performance of FLAME is insensitive to different values of $C$. In addition, it can be shown for any $C$, FLAME is Fisher consistent, by applying the general results in \cite{Lin2004note}. Thus, the default value for $C$ as proposed in \cite{marron2007distance} will be used in FLAME. In this section, we introduce two ways of choosing the second parameter $\theta$. As the property and the performance of FLAME depends on the choice of this parameter, it is important to select the right amount of thresholding. In the following two subsections, we discuss two options of choosing parameter $\theta$. The first option is based on empirical plots resulting from the training data and is of practical useful, and it is the $\theta$ value that we suggest. The second option is motivated by a theoretical consideration and is heuristically meaningful as well.

\subsection{Parameter from performance trade-off}\label{sec:suggesttheta}
Note that, an optimal $\theta$ depends on the nature of the data and problems that users have. The optimal $\theta$ also depends on two performance measures: insensitive to the imbalanced data, and resistance to overfitting. However, without prior knowledge, we may want to have a ``good'' trade-off between them. In this subsection, we suggest the following method to choose $\theta$ if no further information is provided.

As will become clear from the simulation examples shown in Section \ref{simulation2}, we have observed that several performance measures for the FLAME classifiers, for example, the within-group error (see the definition in Section \ref{sec:simulations} and also in \cite{qiao2009adaptive}), are monotonically decreasing with respect to $\theta$. On the other hand, performance measures such as the \textit{RankComp} (see also in Section \ref{sec:simulations}), are monotonically increasing functions of $\theta$. The RankComp measure is more related to the overfitting phenomena, and the within-group error is designed for measuring the performance against the imbalanced data. The lesson is that with $\theta$ increases, FLAME becomes less sensitive to the imbalanced data issue, but is subject to more overfitting. This motivates us to use the following parameter: the two curves of the two measures are normalized to be between 0 and 1. When $\theta=0$, the FLAME classifier (equivalent to DWD)  has the smallest RankComp measure 0, but the largest within-group error 1. When $\theta=1$, the FLAME classifier (equivalent to SVM) has the smallest within-group error 0, but the largest RankComp 1. The suggested $\theta$ is chosen as the value where the two normalized curves intersect, that is, the normalized within-group error is the same as the normalized RankComp for this $\theta$. This suggested parameter represents a natural trade-off between the two measures: neither measure is absolutely optimal, but each measure compromises by the same relative amount. This suggested parameter is called the  \textit{equal-trade-off parameter}.  

\subsection{Parameter from adaptive modeling}\label{sec:adaptivetheta}
Having observed that the DWD discrimination direction is usually closer to the Bayes rule direction, but its location term $\beta$ is sensitive to the imbalanced data issue, we propose the following alternative data-driven approach to select an appropriate $\theta$. Without loss of generality, we assume that the negative class is the majority class with sample size $n_-$ and the positive class is the minority class with sample size $n_+$. We point out that the main reason that DWD is sensitive to the imbalanced data issue is that it uses all vectors in the \textit{majority} class to build up a classifier. A heuristic strategy to correct this would be to force the optimization to use the same number of vectors from both classes to build up a classifier: we first apply DWD to the data set, and calculate the distances of all data in the majority class to the current DWD classification boundary; we then train FLAME with a carefully chosen parameter $\theta$ which assigns positive loss to the closet $n_+$ data vectors in the majority class to the classification boundary. As a consequence, each class will have exactly $n_+$ vectors which have positive loss. In other words, while keeping the least imbalance (because we have the same numbers of vectors from both classes that have influence over the optimization), we obtain a model with the least possible overfitting (because $2n_+$ vectors have influence, instead of only the limited support vectors as in SVM.)

In practice, since the new FLAME classification boundary using the $\theta$  chosen above may be different from the initial DWD classification boundary, the $n_+$ closest points to the FLAME classification boundary may not be the same $n_+$  closest points to the DWD boundary. This means that it is not guaranteed that exactly $n_+$ points from the majority class will have positive loss. However, one can expect that reasonable approximation can be achieved. Moreover, an iterative scheme for finding $\theta$ is introduced as follows in order to minimize such discrepancy.

For simplicity, we let $(\xv_i,y_i)$ with index $i$ be an observation from the positive/minority class and $(\xv_j,y_j)$ with index $j$ be an observation from the negative/majority class.
\begin{algorithm}\label{algorithm1} (\textbf{Adaptive parameter})
    \begin{enumerate}
	\item Initiate $\theta_0=0$.
	\item For $k=0,1,\cdots$,
	    \begin{enumerate}
		\item Solve FLAME solutions $\omegav(\theta_k)$ and $\beta(\theta_k)$ given parameter $\theta_k$.
		\item Let $\ds{\theta_{k+1}=\max\left(\theta_k,~\left\{g_{(n_+)}(\theta_k)\sqrt{C}\right\}^{-1}\right)}$, where $g_j(\theta_k)$ is the functional margin $u_j\equiv y_{j}(\xv_{j}^T\omegav(\theta_k)+\beta(\theta_k))$ of the $j$th vector in the negative/majority class and $g_{(l)}(\theta_k)$ is the $l$th order statistic of these functional margins.
	    \end{enumerate}
	\item When $\theta_k = \theta_{k-1}$, the iteration stops.
    \end{enumerate}
\end{algorithm}
The goal of this algorithm is to make $g_{(n_+)}(\theta_k)$ to be the greatest functional margin among all the data vectors that have positive loss in the negative/majority class. To achieve this, we calibrate $\theta$ by aligning $g_{(n_+)}(\theta_k)$ to the turning point $u=1/(\theta\sqrt{C})$ in the definition of the FLAME loss (\ref{turning}), that is $g_{(n_+)}(\theta_k)=1/(\theta\sqrt{C})\Rightarrow\theta = \left(g_{(n_+)}(\theta_k)\sqrt{C}\right)^{-1}$.

We define the equivalent sample objective function of FLAME for the iterative algorithm above,
$\ds
 s(\omegav,\beta,\theta) =
\frac{1}{n_++n_-} \left[\sum_{i=1}^{n_+} L((\xv_i^T\omegav+\beta),\theta)+\sum_{j=1}^{n_-} L(-(\xv_j'\omegav+\beta),\theta)\right] + \frac{\lambda}{2}\|\omegav\|^2.
$
Then the convergence of this algorithm is shown in Theorem \ref{thm1:algorithm_converge}.
\begin{theorem}\label{thm1:algorithm_converge}
In Algorithm \ref{algorithm1}, $s(\omegav_k,\beta_k,\theta_k)$ is non-increasing in $k$. As
a consequence, Algorithm \ref{algorithm1} converges to a stationary point
$s(\omegav_\infty,\beta_\infty,\theta_\infty)$ where $s(\omegav_k,\beta_k,\theta_k)\geq
s(\omegav_\infty,\beta_\infty,\theta_\infty)$. Moreover, Algorithm \ref{algorithm1} terminates
finitely.
\end{theorem}
Ideally, one would hope to get an optimal parameter $\theta^{*}$ which satisfies 
$\ds{\theta^{*}=\left(g_{(n_+)}(\theta^{*})\sqrt{C}\right)^{-1}}.$
In practice, $\theta_{\infty}$ will approximate $\theta^{*}$ very well. In addition, we notice that one-step iteration usually gives decent results for simulation examples and some real examples.

\section{Theoretical Properties}\label{sec:theory}
In this section, several important theoretical properties of the FLAME classifiers are investigated. We first prove the Fisher consistency \citep{Lin2004note} of the FLAME in Section \ref{sec:fisher}. As one focus of this paper is imbalanced data classification, the asymptotic properties for FLAME under extremely imbalanced data setting is studied in Section \ref{sec:d_fix_m_large}. Lastly, a novel HDLSS asymptotics where $n$ is fixed and $d\rightarrow \infty$, the other focus of this article, is studied in Section \ref{sec:hdlss_theory}.

\subsection{Fisher consistency}\label{sec:fisher}
Fisher consistency is a very basic property for a classifier. A classifier is Fisher consistent means that the minimizer of the conditional risk of the classifier given observation $\xv$ has the same sign as the Bayes rule, ${\ds\argmax_{k\in\set{+1,-1}}} \Pr(Y=k|\Xv=\xv)$. It has been shown that both SVM and DWD are Fisher consistent \citep{Lin2004note, Qiao2010Weighteda}. The following proposition states that the FLAME classifiers are Fisher consistent too.

\begin{prop}\label{prop2:fisher}
 Let $f^*$ be the global minimizer of $E[L(Yf(\Xv),\theta)]$, where $L(\cdot,\theta)$
is the loss function for FLAME given parameter $\theta$. Then $\sgn\left(f^*(\xv)\right)
=\sgn\left(\Pr(Y=+1|\Xv=\xv)-1/2\right)$.
\end{prop}

\subsection{Asymptotics under imbalanced setting}\label{sec:d_fix_m_large}
In this subsection, we investigate the asymptotic performance of SVM, DWD and FLAME. The asymptotic setting we focus on is when the minority sample size $n_+$ is fixed and the majority sample size $n_-\rightarrow \infty$, which is similar to the setting in \cite{owen2007infinitely}. We will show that DWD is sensitive to the imbalanced data, while FLAME with proper choices of parameter $\theta$ and SVM are not.

Let $\overline\xv_+$ be the sample mean of the positive/minority class. Theorem \ref{thm3:oDWD_intercept_diverge} shows that in the imbalanced data setting, when the size of the negative/majority class grows while that of the positive/minority class is fixed, the intercept term for DWD tends to negative infinity, in the order of $\sqrt{m}$. Therefore, DWD will classify all the observations to the negative/majority class, that is, the minority class will be $100\%$ misclassified.

\begin{theorem}\label{thm3:oDWD_intercept_diverge}
Let $n_+$ be fixed. Assume that the conditional distribution of the negative majority class $F_-(\xv)$ surrounds $\overline\xv_+$ by the definition given in \citet{owen2007infinitely}, and that $\gamma$ is a constant satisfying $\ds\inf_{\|\omegav\|=1}\int_{(\xv-\overline\xv_+)'\omegav>0}dF_-(\xv)>\gamma\geq 0$, then the
DWD intercept $\wh\betav$ satisfies 
\[
\wh\betav<-\sqrt{\frac{\gamma}{C}m}-\overline\xv_+^T\omegav=-\sqrt{\frac{n_-\gamma}{n_+C}}-\overline\xv_+^T\omegav.
\]
\end{theorem}

In Section \ref{sec:adaptivetheta}, we have introduced an iterative approach to select the parameter $\theta$. Theorem \ref{thm4:w_beta_FLAME} shows that with the optimal parameter $\theta^*$ found by Algorithm \ref{algorithm1}, the discriminant direction of FLAME is in the same direction of the vector that joins the sample mean of the positive class and the \textit{tilted} population mean of the negative class. Moreover, in contrast to DWD, the intercept term of FLAME in this case is finite.

\begin{theorem}\label{thm4:w_beta_FLAME}
Suppose that $n_-\gg n_+$ and $\omegav^*$ and $\beta^*$ are the FLAME solutions trained with the parameter $\theta^*$ that satisfies $\ds{\theta^{*}=\left({g_{(n_+)}(\theta^{*})\sqrt{C}}\right)^{-1}}$. Then $\omegav^*$ and $\beta^*$ satisfy that
\begin{align*}
 \omegav^*
&=\frac{C}{(1+m)\lambda}\left[\overline\xv_+-\frac{\int
(\xv^T\omegav^*+\beta^*)^{-2}\xv
dF_-(\xv\mid E)}{\int(\xv^T\omegav^*+\beta^*)^{-2}dF_-(\xv\mid E)}\right],
\end{align*}
where $E$ is the event that
$[Y(\Xv^T\omegav^*+\beta^*)]^{-1}\geq\theta^*\sqrt{C}$ where $(\Xv,Y)$ is a random sample from the negative/majority class, and that
\begin{align*}
	\int(\xv^T\omegav^*+\beta^*)^{-2}dF_-(\xv\mid E)=\frac{n_+}{n^o}C, \mb{where }0<n^o\leq n_+.
\end{align*}
\end{theorem}

As a consequence of Theorem \ref{thm4:w_beta_FLAME}, when $m={n_-}/{n_+}\rightarrow \infty$, we have $\|\omegav^*\|\rightarrow 0$. Since the right-hand-side of the last equation above is positive finite, $\beta^*$ does not diverge. In addition, since $\Pr(\overline E)\rightarrow 1$ with probability converging to 1, $\beta^*<-1/(\theta\sqrt{C})$.

The following theorem shows the performance of SVM under the imbalanced data context, which completes our comparisons between SVM, DWD and FLAME.  
\begin{theorem}\label{thm5:w_beta_SVM}
Suppose that $n_-\gg n_+$. The solutions $\wh\omegav$ and $\wh\beta$ to SVM satisfy that
\begin{align*}
	\wh\omegav=\frac{1}{(1+m)\lambda}\left\{\overline{\xv}_+-\int \xv dF_-(\xv\mid G)\right\},
\end{align*}
where $G$ is the event that $1-Y(\Xv^T\wh\omegav+\wh\beta)>0$ where $(\Xv,Y)$ is a random sample from the negative/majority class, and that
\begin{align*}
	\Pr(\overline G)=\Pr(1+\Xv^T\wh\omegav+\wh\beta\leq 0)=1-1/m.
\end{align*}
\end{theorem}


The last statement in Theorem \ref{thm5:w_beta_SVM} means that with probability converging to 1,  $\wh\beta\leq -1$. However, note this is the only restriction that SVM solution has for the intercept term (recall that the counterpart in DWD is $\wh\betav<-\sqrt{\frac{\gamma}{C}m}-\overline\xv_+^T\omegav$).

\subsection{High-dimensional, low-sample size asymptotics}\label{sec:hdlss_theory}
HDLSS data are emerging in many areas of scientific research. The HDLSS asymptotics is a recently developed theoretical framework. \citet{hall2005geometric} gave a geometric representation for the HDLSS data, which can be used to study these new `$n$ fixed, $d\rightarrow\infty$' asymptotic properties of binary classifiers such as SVM and DWD. \citet{ahn2007high} weakened the conditions under which the representation holds. \citet{Qiao2010Weighteda} improved the conditions and applied this representation to investigate the performance of the weighted DWD classifier. The same geometric representation can be used to analyze FLAME. See summary of some previous HDLSS results in the online supplementary materials. We develop the HDLSS asymptotic properties of the FLAME family by providing conditions in Theorem \ref{thm6:flame_hdlss_cutoff} under which the FLAME classifiers always correctly classify HDLSS data.

We first introduce the notations and give some regularity assumptions, then state the main theorem. Let $k\in\set{+1,-1}$ be the class index. For the $k$th class and given a fixed $n_k$, consider a sequence of random data matrices $\Xv^k_1,\Xv^k_2,\cdots \Xv_d^k,\cdots$, indexed by the number of rows $d$, where each column of $\Xv_d^k$ is a random observation vector from $\R^d$ and each row represents a variable. Assume that each column of $\Xv_d^k$ comes from a multivariate distribution with dimension $d$ and with covariance matrix $\Sigmav_d^k$ independently. Let $\lambda^k_{1,d}\geq\cdots\geq\lambda^k_{d,d}$ be the eigenvalues of the covariance, and ${\left(\sigma_d^k\right)}^2=d^{-1}\sum_{i=1}^d\lambda_{i,d}^k$ the average eigenvalue. The eigenvalue decomposition of $\Sigmav_d^k$ is $\Sigmav_d^k=\Vv_d^k\Lambdav_d^k {\left(\Vv_d^k\right)}^T$. We may define the square root of $\Sigmav_d^k$ as ${\left(\Sigmav_d^k\right)}^{1/2}=\Vv_d^k{\left(\Lambdav_d^k\right)}^{1/2}$, and the inverse square root ${\left(\Sigmav_d^k\right)}^{-1/2}=\left(\Lambdav_d^k\right)^{-1/2}\left(\Vv_d^k\right)^T$. With minimal abuse of notation, let $\E(\Xv_d^k)$ denote the expectation of columns of $\Xv_d^k$. Lastly, the $n^k\times n^k$ \textit{dual} sample covariance matrix is denoted by $\Sv^k_{D,d} = d^{-1}{\left\{\Xv_d^k-\E(\Xv_d^k)\right\}}^T\left\{\Xv_d^k-\E(\Xv_d^k)\right\}$. 

\begin{assumption}\label{assum1}
There are five components:
		\begin{enumerate}[label=(\roman{*})]
				\item Each column of $\Xv_d^k$ has mean $\E(\Xv_d^k)$ and the covariance matrix $\Sigmav_d^k$ of its distribution is positive definite.
				\item The entries of
				$\Zv_d^k\equiv{\left(\Sigmav_d^k\right)}^{-\frac{1}{2}}\left\{\Xv_d^k-\E(\Xv_d^k)\right\}=\left(\Lambdav_d^k\right)^{-\frac{1}{2}}\left(\Vv_d^k\right)^T\left\{\Xv_d^k-\E(\Xv_d^k)\right\}$ are independent.
		
				\item The fourth moment of each entry of each column is uniformly bounded by
				$M>0$ and the Wishart representation holds for each dual sample covariance matrix $\Sv^k_{D,d}$ associated with $\Xv_d^k$, that is,	
				\begin{align*}
					d\Sv_{D,d}^k&=	\left\{{\left(\Zv_d^k\right)}^T{\left(\Lambdav_d^k\right)}^{1/2}{\left(\Vv_d^k\right)}^T\right\}\left\{\Vv_d^k{\left(\Lambdav_d^k\right)}^{1/2}\Zv_d^k\right\}=\sum_{i=1}^d\lambda^k_{i,d}\Wv_{i,d}^k,
				\end{align*}
					where $\Wv_{i,d}^k\equiv {\left(Z_{i,d}^k\right)}^T Z_{i,d}^k$ and $Z_{i,d}$ is the $i$th row of $\Zv_d^k$ defined above. It is called Wishart representation because if $\Xv_d^k$ is Gaussian, then each $\Wv_{i,d}^k$ follows the Wishart distribution
		$\Wc_{n^k}(1,\Id_{n^k})$ independently.
				\item The eigenvalues of $\Sigmav_d^k$ are sufficiently diffused, in the sense that
				\begin{equation}\label{eqn:assum1}
				\epsilon_d^k=\frac{\sum_{i=1}^d(\lambda_{i,d}^k)^2}{(\sum_{i=1}^d\lambda^k_{i,d})^2}\rightarrow 0~~\mb{as}~~d\rightarrow \infty.
				\end{equation}
				\item The sum of the eigenvalues of  $\Sigmav_d^k$ is the same order as $d$, in the
				sense that ${\left(\sigma_d^k\right)}^2=O(1)$ and $1/{\left(\sigma_d^k\right)}^2=O(1)$.
		\end{enumerate}
\end{assumption}

\begin{assumption}\label{assum2}
The distance between the two population expectations satisfies,
\begin{align*}
  d^{-1}\big\|\E(\Xv_d^{(+1)})-\E(\Xv_d^{(-1)})\big\|^2\rightarrow\mu^2,~\mb{as}~d\rightarrow\infty.
\end{align*}
Moreover, there exist constants $\sigma^2$ and $\tau^2$, such that
$${\left(\sigma_d^{(+1)}\right)}^2\rightarrow \sigma^2,\mb{ and }{\left(\sigma_d^{(-1)}\right)}^2\rightarrow \tau^2.$$
\end{assumption}

Let $\nu^2\equiv \mu^2+\sigma^2/n_++\tau^2/n_-$. The following theorem gives the sure classification condition for FLAME, which includes  SVM and DWD as special cases.
\begin{theorem}\label{thm6:flame_hdlss_cutoff}
Without loss of generality, assume that $n_+\leq n_-$. The situation of $n+>n_-$ is similar and omitted.
	\begin{itemize}
		\item If either one of the following three conditions is satisfied,
			\begin{enumerate}
				\item for $\theta\in\left[0,(1+\sqrt{m^{-1}})/(\nu\sqrt{dC})\right)$, $\mu^2>(n_-/n_+)^{\frac{1}{2}}\sigma^2/n_+-\tau^2/n_->0$;
				\item for $\theta\in\left[(1+\sqrt{m^{-1}})/(\nu\sqrt{dC}),2/(\nu\sqrt{dC}) \right)$, $\mu^2>T-\tau^2/n_->0$ where $T:=\left(1/(2\theta\sqrt{dC})+\sqrt{1/(4\theta^2dC)+\sigma^2/n_+}\right)^2-\sigma^2/n_+$;
				\item for $\theta\in\left[2/(\nu\sqrt{dC}),1\right]$, $\mu^2>\sigma^2/n_+-\tau^2/n_->0$,
			\end{enumerate}
			then for a new data point $\xv^+_0$ from the positive class ($+1$), \\
			$\Pr(\xv^+_0~~\mb{is correctly classified by FLAME})\rightarrow 1,~\mb{as}~d\rightarrow \infty.$\\
			Otherwise, the probability above $\rightarrow 0$.
		\item If either one of the following three conditions is satisfied,
			\begin{enumerate}
				\item for $\theta\in\left[0,(1+\sqrt{m^{-1}})/(\nu\sqrt{dC})\right)$, $(n_-/n_+)^{\frac{1}{2}}\sigma^2/n_+-\tau^2/n_->0$;
				\item for $\theta\in\left[(1+\sqrt{m^{-1}})/(\nu\sqrt{dC}),2/(\nu\sqrt{dC}) \right)$, $T-\tau^2/n_->0$;
				\item for $\theta\in\left[2/(\nu\sqrt{dC}),1\right]$, $\sigma^2/n_+-\tau^2/n_->0$,
			\end{enumerate}
			then for any $\mu>0$, for a new data point $\xv^-_0$ from the negative class ($-1$),\\
		$\Pr(\xv^-_0~~\mb{is correctly classified by FLAME})\rightarrow 1,~\mb{as}~d\rightarrow
		\infty.$
	\end{itemize}
\end{theorem}

Theorem \ref{thm6:flame_hdlss_cutoff} has two parts. The first part gives the conditions under which FLAME correctly classifies a new data point from the positive class, and the second part is for the negative class. Each part lists three conditions based on three disjoint intervals of parameter $\theta$. Note the first and third intervals of each part generalize results which were shown to hold only for DWD and SVM before \citep[\textit{c.f.} Theorem 1 and Theorem 2 in][]{hall2005geometric}. In particular, it shows that all the FLAME classifiers with $\theta$ falling into the first interval behave like DWD asymptotically. Similarly, all the FLAME classifiers with $\theta$ falling into the third interval behave like SVM asymptotically. This partially explains the shape of the within-group error curve that we will show in Figures \ref{fig:tryoutInterchangeable}, \ref{fig:tryoutIndependent}, and \ref{fig:tryoutBlockCorrelated}, which we will discuss in the next section.

In the first part, the condition for other FLAMEs (with $\theta$ in the second interval) is weaker than the DWD-like FLAMEs (in the first interval), but stronger than the SVM-like FLAMEs (in the third interval). This means that it is easier to classify a new data point from the positive/minority class by SVM, than by an intermediate FLAME, which is easier than by DWD. Note that when $n_+\leq n_-$, the hyperplane for FLAME is in general closer to the positive class.

In terms of classifying data points from the negative class, the order of the difficulties among DWD, FLAME and SVM reverses.

\section{Simulations}\label{sec:simulations}
FLAME is not only a unified representation of DWD and SVM, but also introduces a new family of classifiers which are capable of avoiding the overfitting HDLSS data issue and the sensitivity to imbalanced data issue. In this section, we use simulations to show the performance of FLAME at various parameter levels. We will show that with a range of carefully chosen parameters, FLAME can outperform both the DWD and the SVM methods in various simulation settings.

\subsection{Measures of performance}
Before we introduce our simulation examples, we first introduce the performance measures in this paper. Note that the Bayes rule classifier can be viewed as the ``gold standard'' classifier. In our simulation settings, we assume that data are generated from two Gaussian populations $MVN(\muv_\pm,\Sigmav)$ with different mean vectors $\muv_+$ and $\muv_-$ and same covariance matrices $\Sigmav$. This setting leads to the following Bayes rule.
\begin{align}\label{bayesrule}
	\sgn(\xv^T\omegav_B+\beta_B)\mb{ where }\omegav_B = \Sigmav^{-1}(\muv_+-\muv_-)\mb{ and }\beta_B=-\half(\muv_++\muv_-)'\omegav_B.
\end{align}

Five performance measures are evaluated in this paper:
\begin{enumerate}
		\item The mean within-class error (MWE) for out-of-sample test set, which is defined as
\begin{align*}
MWE & =\frac{1}{2n_+}\sum_{i=1}^{n_+} \mathbbm{1}(\widehat{Y}^+_{i} \neq Y^+_{i}) +
\frac{1}{2n_-}\sum_{j=1}^{n_-} \mathbbm{1}(\widehat{Y}^-_{j} \neq Y^-_{j})
\end{align*}
		\item The deviation of the estimated intercept $\beta$ from the Bayes rule intercept $\beta_B$: $|\beta-\beta_B|$.
		\item Dispersion: a measure of the stochastic variability of the estimated discrimination direction vector $\omegav$. The dispersion measure was introduced in Section \ref{sec:intro}, as the trace of the sample covariance of the resulting discriminant direction vectors: $\mb{disperson} = \var([\omegav_r]_{r=1:R})$ where $R$ is the number of repeated runs.
		\item Angle between the estimated discrimination direction $\omegav$ and the Bayes rule direction $\omegav_B$: $\angle(\omegav,\omegav_B)$.
		\item RankComp$(\omegav,\omegav_B)$: In general, for two direction vectors $\omegav$ and $\omegav^*$, RankComp is defined as the proportion of the pairs of variables, among all $d(d-1)/2$ pairs, whose relative importances (in terms of their absolute values) given by the two directions are different, \ie,
$$\mathrm{RankComp}(\omegav,\omegav^*)\equiv\frac{1}{d(d-1)/2}\sum_{1\leq i<j\leq d} \mathbbm{1}\left\{(|\omega_i|-|\omega_j|)\times(|\omega^*_i|-|\omega^*_j|)<0\right\},$$
		where $\omega_i$ and $\omega_i^*$ are the $i$th components of the vectors $\omegav$ and $\omegav^*$ respectively.
		
		We report the RankComp between the estimated direction $\omegav$ and the Bayes rule direction $\omegav_B$ to measure their closeness.
\end{enumerate}
We will investigate these measures based on different dimensions $d$ and different imbalance factors $m$.

\begin{figure}[!ht]
  \begin{center}
   \includegraphics[height=1\linewidth, width=0.7\linewidth, angle=270]{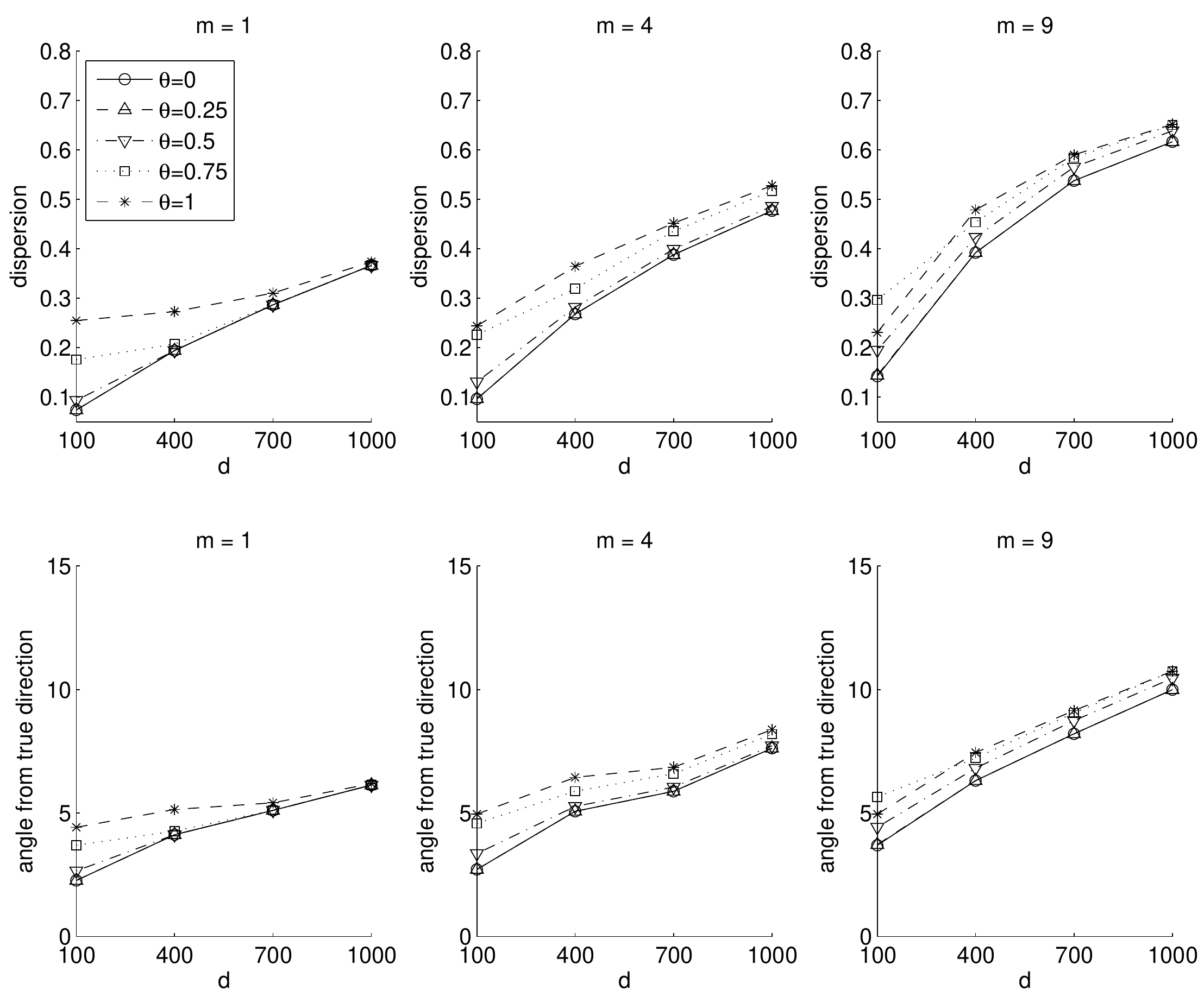}
  \end{center}
  \caption{The dispersions (top row) and the angles between the FLAME direction and the Bayes direction (bottom row) for 50 runs of simulations, where the imbalance factors $m$ are 1, 4 and 9 (the left, center and right panels), in the increasing dimension setting ($d=100, 400, 700, 1000$; shown on the $x$-axes). The FLAME machines have $\theta=0, 0.25, 0.5, 0.75, 1$ which are depicted using different curve styles (the first and the last cases correspond to DWD and SVM, respectively.) Note that with $\theta$ and the dimension $d$ increase, both the dispersion and the deviation from the Bayes direction increase. The emergence of the imbalanced data (the increase of $m$) does not much deteriorate the FLAME directions except for large $d$.}
 \label{fig:HD_disp_angle}
\end{figure}
\subsection{Effects of dimensions and imbalanced data}
In Section \ref{sec:intro}, a specific FLAME ($\theta=.5$) has been compared with SVM ($\theta=1$) and DWD ($\theta=0$) in Figure \ref{fig:unit_sphere}, and on average, its discriminant directions are closer to the Bayes rule direction $\omegav_B$ compared to the SVM directions, but is less close than the DWD directions. In this subsection, we will further investigate the performance of FLAME with several different values of $\theta$, and compare them with DWD and SVM under various simulation settings.

Figure \ref{fig:HD_disp_angle} shows the comparison results under the same simulation setting with various combinations of $(d, m)$'s. In this simulation setting, data are from multivariate normal distributions with identity covariance matrices $MVN_d(\muv_\pm,\Id_d)$, where $d= 100, 400, 700\mb{ and }1000$. We let $\muv_0 = c(d,d-1,d-2,\cdots,1)^T$ where $c>0$ is a constant which scales $\muv_0$ to have norm 2.7. Then we let $\muv_+=\muv_0$ and $\muv_-=-\muv_0$. The imbalance factor varies among 1, 4 and 9 while the total sample size is 240. For each experiment, we repeat the simulation 50 times, and plot the average performance measure in Figure \ref{fig:HD_disp_angle}. The Bayes rule is calculated according to (\ref{bayesrule}). It is obvious that when the dimension increases, both the dispersion and the angle increase. They are indicators of overfitting HDLSS data. When the imbalance factor $m$ increases, the two measures increases as well, although not as much as when the dimension increases. More importantly, it shows that when $\theta$ decreases (from 1 to 0, or equivalently FLAME changes from SVM to DWD), the dispersion and the angle both decrease, which is promising because it shows that FLAME improves SVM in terms of the overfitting issue.

\subsection{Effects of tuning parameters with covariance}\label{simulation2}
We also investigate the effect of different covariance structures, since independence structure among variables as in the last subsection is not realistic in real applications. We investigate three covariance structures: independent, interchangeable and block-interchangeable covariance. Data are generated from two multivariate normal distributions $MVN_{300}(\muv_\pm,\Sigmav)$ with $d=300$. We fist let $\muv_1=(75,74,73,\cdots,1,0,0,\cdots,0)'$, then scale it by multiply a constant $c$ such that the Mahalanobis distance between $\muv_+=c\muv_1$ and $\muv_-=-c\muv_1$ equals 5.4, \ie, $\ds{(\muv_+-\muv_-)'\Sigmav^{-1}(\muv_+-\muv_-)=5.4}$. Note that this represents a reasonable signal-to-noise ratio.

We consider the FLAME machines with different parameter $\theta$ from a grid of 11 values $(0,0.1,0.2,\cdots,1)$, and apply them to nine simulated examples (three different imbalance factors ($m=2, 3, 4$) $\times$ three covariance structures). For the independent structure example, $\Sigmav=\Id_{300}$; For the interchangeable structure example, $\Sigmav_{ii}=1$ and $\Sigmav_{ij}=0.8$ for $i\neq j$; For the block-interchangeable structure example, we let $\Sigmav$ be a block diagonal matrix with five diagonal blocks, the sizes of which are 150, 100, 25, 15, 10, and each block is an interchangeable covariance matrix with diagonal entries 1 and off-diagonal entries 0.8.

\begin{figure}[!ht]
  \begin{center}
   \includegraphics[height=1\linewidth, width=0.7\linewidth, angle=270]{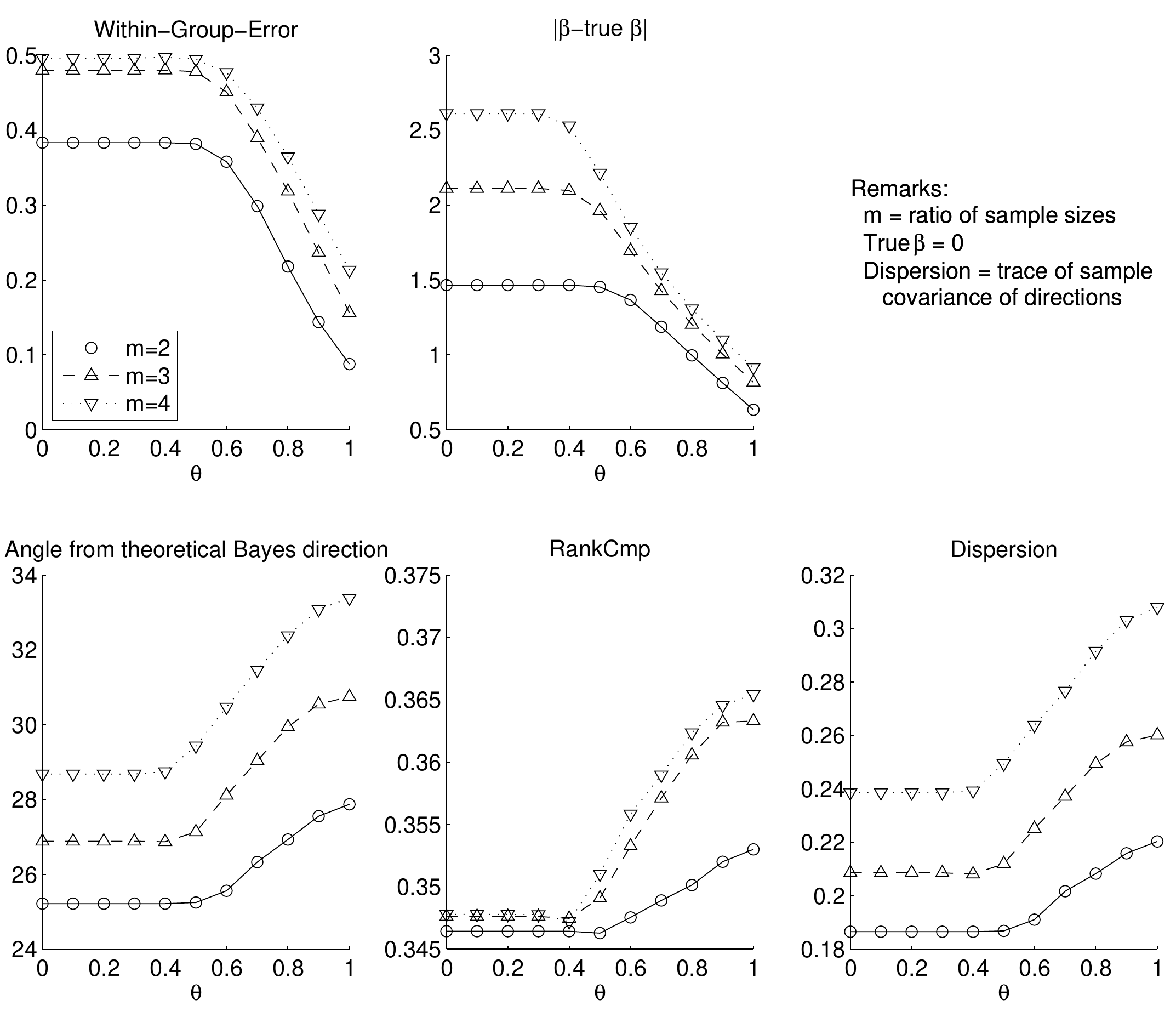}
  \end{center}
  \caption{Interchangeable example. It can be seen that with FLAME turns from DWD to SVM ($\theta$ from 0 to 1), the within-class error decreases (top-left), thanks to the more accurate estimate of the intercept term (top-middle). On the other hand, this comes at the cost of larger deviation from the Bayes direction (bottom-left), incorrect rank of the importance of the variables (bottom-middle) and larger stochastic variability of the estimation directions (bottom-right).}
 \label{fig:tryoutInterchangeable}
\end{figure}

Figure \ref{fig:tryoutInterchangeable} provides the summary results of the interchangeable structure example. Since the results are similar under different covariance structures, results from the other two covariance structures are included in the online supplementary materials to save space (Figure \ref{fig:tryoutIndependent} for the independent structure, and Figure \ref{fig:tryoutBlockCorrelated} for the block-interchangeable covariance).

In each plot, we include the within-group error (top-left), the absolute value of the difference between the estimated intercept and the Bayes intercept $|\beta-\beta_B|$ (top-middle), the angle between the estimated direction and the Bayes direction $\angle(\omegav,\omegav_B)$ (bottom-left), the RankComp between the estimated direction and the Bayes direction (bottom-middle) and the dispersion of the estimated directions (bottom-right).

We can see that in Figure \ref{fig:tryoutInterchangeable} (and Figures \ref{fig:tryoutIndependent} and \ref{fig:tryoutBlockCorrelated} in the online supplementary materials), when we increase $\theta$ from 0 to 1, \ie, when the FLAME moves from the DWD end to the SVM end, the within-group error decreases. This is mostly due to the fact that the intercept term $\beta$ comes closer to the Bayes rule intercept $\beta_B$. On the other hand, the estimated direction is deviating from the true direction (larger angle), is giving the wrong rank of the variables (larger RankComp) and is more unstable (larger dispersion). Similar observations hold for the other two covariance structures, with one exception in the block interchangeable setting (Figure \ref{fig:tryoutBlockCorrelated}) where the RankComp first decreases then increases.

In the entire FLAME family, DWD represents one extreme which provides better estimation of the direction, is closer to the Bayes direction, provides the right order for all variables, and is more stable. But it suffers from the inaccurate estimation of $\beta$ in the presence of imbalanced data; SVM represents the other extreme, which is not sensible to imbalanced data and usually provides a good estimation of $\beta$, but is in general outperformed by DWD in terms of closeness to the Bayes optimal direction. In most situations, within the FLAME family, there is no single machine that is better than the both ends from the two aspects at the same time.

The observations above motivate the use of the equal-trade-off parameter introduced in Section \ref{sec:suggesttheta}. In the next subsection, we will compare this parameter choice with other options.

\subsection{Comparison of parameter options}\label{sec:sim_parameter}
\begin{figure}[!ht]
  \begin{center}
   \includegraphics[height=0.56\linewidth, width=0.8\linewidth]{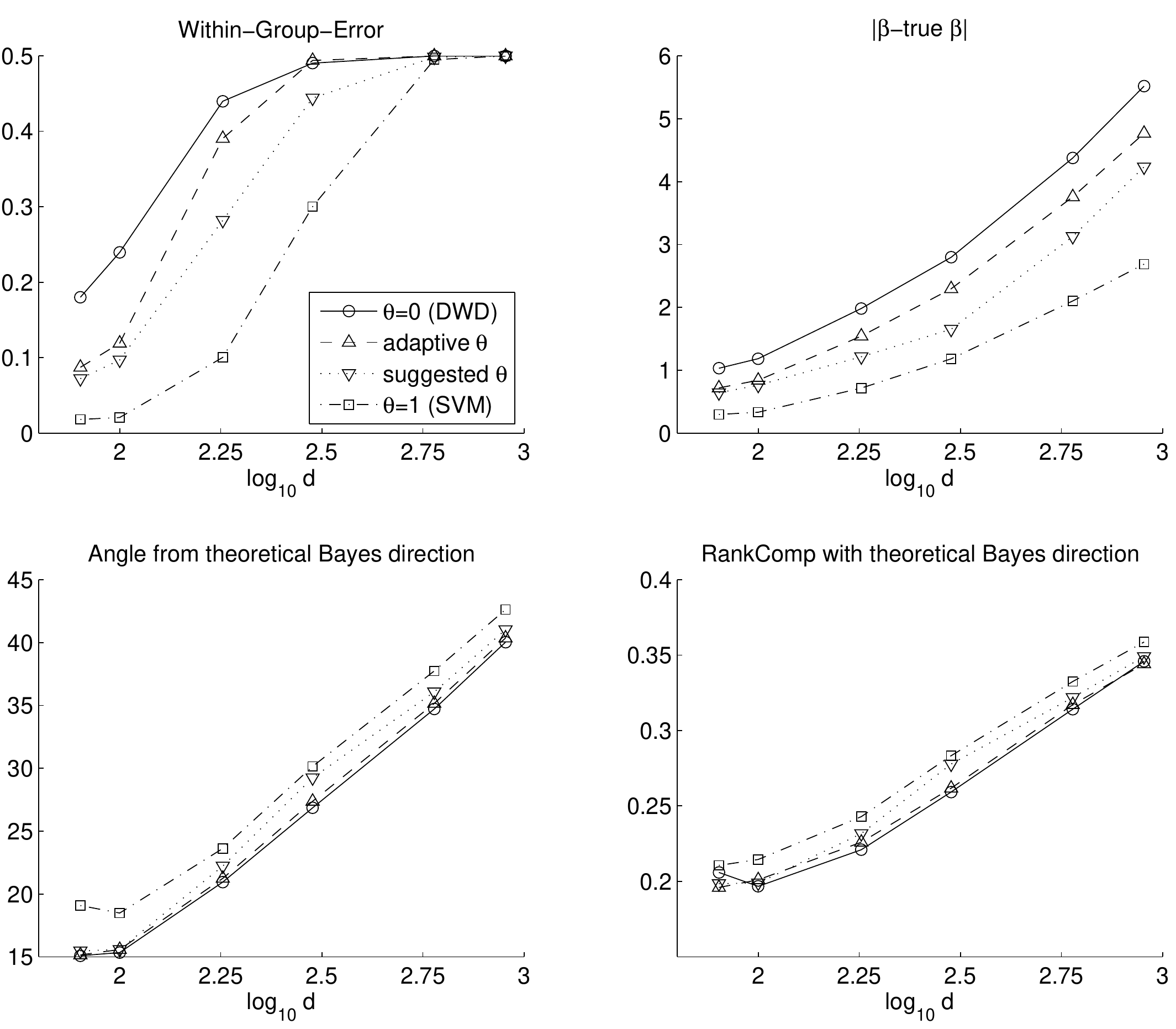}
  \end{center}
  \caption{Comparison of four FLAMEs with $\theta=0,~1$, the suggested $\theta$ introduced in Section \ref{sec:suggesttheta} and the adaptive $\theta$ (after one step) introduced in Section \ref{sec:adaptivetheta} for a simulated example with block-interchangeable dependence structure, in terms of the within-group error, deviation from the true intercept term $\beta_B$, deviation from the true direction $\omegav_B$, and the RankComp from $\omegav_B$. Intermediate FLAMEs provide improvements over DWD for the first two measures and over SVM for the last two measures.}
 \label{fig:simulation_blk_interchangeable}
\end{figure}

We have suggested an equal-trade-off parameter based on the plots of within-group error and the RankComp (see Section \ref{sec:suggesttheta}) and justified the use of an adaptive $\theta$ based on an iterative procedure (see details in Section \ref{sec:adaptivetheta}). Figure \ref{fig:simulation_blk_interchangeable} compares FLAME with these two choices, and with $\theta=0$ (DWD) and 1 (SVM). Various covariance structures (independent, interchangeable and block-interchangeable) are investigated. To save the space, we only show the results for the block-interchangeable dependence structure as this is more realistic in many real applications in genomic science and other applications. Here the full dimensions ($d=80, 100, 180, 300, 600\mb{ or }900$) are divided to three blocks (50\%, 25\% and 25\% of $d$). The total sample size is 240 and the imbalance factor $m$ is 3 (moderate imbalanced).

In Figure \ref{fig:simulation_blk_interchangeable}, we compare the within-group error, $|\beta-\beta_B|$, $\angle(\omegav,\omegav_B)$, and $\mb{RankComp}(\omegav,\omegav_B)$. We see that these intermediate FLAMEs provide improvements over DWD for the first two measures and over SVM for the last two measures. For relatively small $d$, the equal-trade-off $\theta$ value is very similar to the adaptive $\theta$. For large $d$, the adaptive $\theta$ is closer to DWD than the equal-trade-off $\theta$. For very large $d$, all four machines encounter difficulty in classification.

\section{Real Data Application}\label{sec:real_data}
In this section we demonstrate the performance of FLAME on a real example: the Human Lung Carcinomas Microarray Dataset, which has been analyzed earlier in \citet{Bhattacharjee2001Classification}. 

The Human Lung Carcinomas Dataset contains six classes: adenocarcinoma, squamous, pulmonary carcinoid, colon,
normal and small cell carcinoma, with sample sizes of 128, 21, 20, 13, 17 and 6 respectively. \citet{liu2008statistical} used this data as a test set to demonstrate their proposed significance analysis of clustering approach. We combine the first two subclasses and the last four subclasses to form the positive and negative classes respectively. The sample sizes are 149 and 56 with imbalance factor $m=2.66$. The original data contain 12,625 genes. We first filter genes using the ratio of the sample standard deviation and sample mean of each gene and keep 2,530 of them with large ratios \citep{dudoit2002comparison,liu2008statistical}.

We conduct a five-fold cross-validation (CV) to evaluate the within-group error for the two classes over 100 random splits. The RankComp measure is calculated based on the full data set instead of on the samples in a single fold. For each replication, we find a suggested value for $\theta$ that leads to the same normalized value for RankComp and within-group error over a grid of $\theta$ values. The adaptive value for $\theta$ (after one step) is calculated based on the DWD direction using all the samples in the data set.

\begin{figure}[!t]
  \begin{center}
   	\includegraphics[height=0.8\linewidth, width=0.56\linewidth, angle=270]{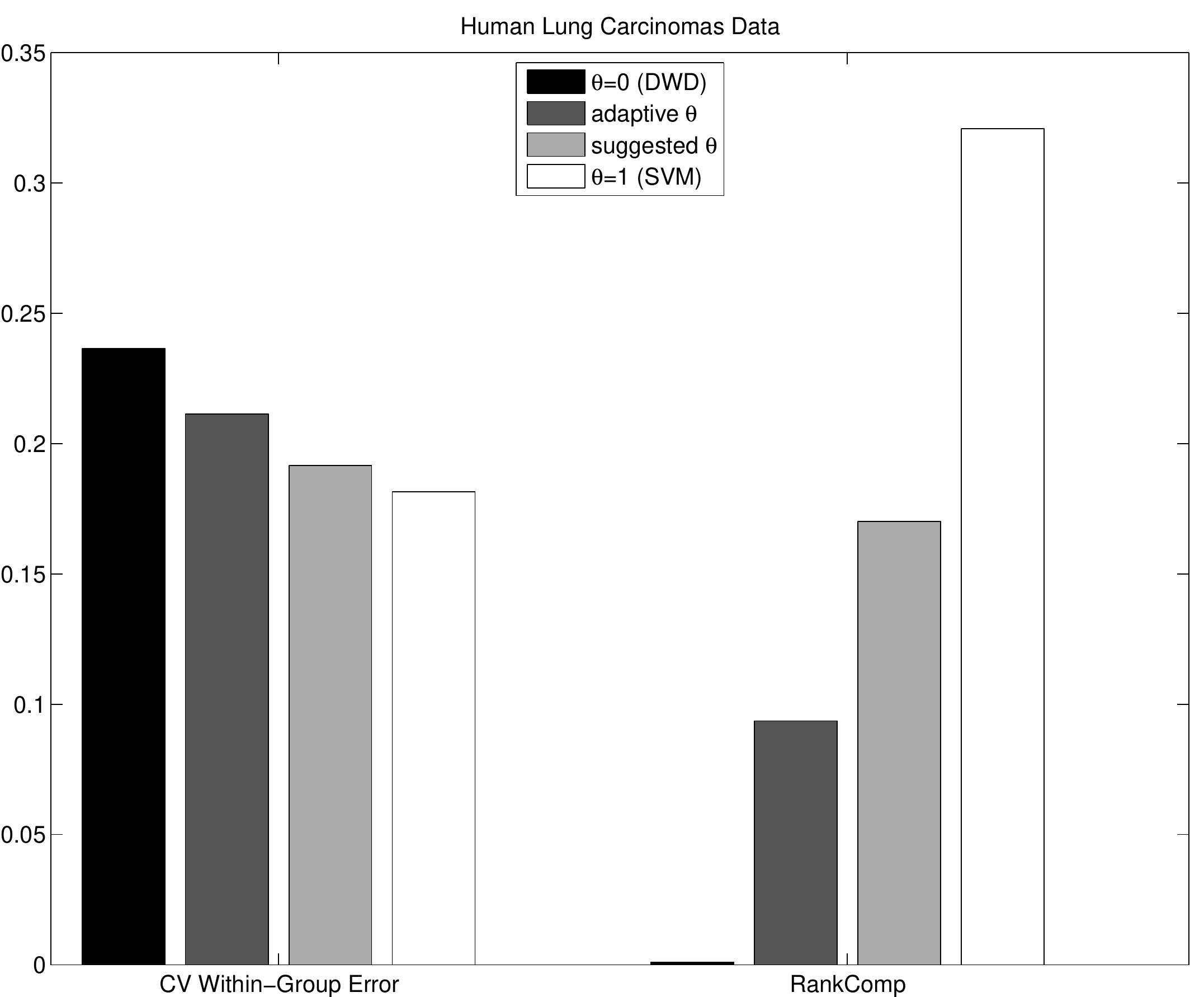}
  \end{center}
  \caption{The average cross-validation within-group errors and the RankComp measures for DWD, FLAME with the adaptive and equal-trade-off $\theta$ parameters and SVM for the Human Lung Carcinomas Dataset.}
  \label{fig:lung1}
\end{figure}

The average within-group errors and the RankComp measures for DWD, SVM, and the two intermediate FLAMEs (using the adaptive $\theta$ and the suggested \textit{equal-trade-off} $\theta$) are shown in Figure \ref{fig:lung1}.

Note that as we do not know the true rank of importance of genes for this real data application (no Bayes rule direction or the rank of importance that it implies), we use the DWD rank as a surrogate to the truth, since simulation examples show that on average its estimated direction is the closest to the Bayes rule direction. Therefore, the RankComp measure for DWD is 0.

This experiment does show that FLAME opens a new dimension of improving both the classification performance and the interpretative ability of the classifier. The compromise of the FLAME classifier, with the suggested equal-trade-off $\theta$, in terms of the within-group error is very small compared to the improvement obtained in terms of the direction. We would not recommend using either DWD or SVM for conducting prediction and interpretation simultaneously due to their bad performance for at least one criterion. A FLAME classifier with an appropriate parameter may be more suitable for practical use.

\section{Conclusion and Discussion}\label{sec:conclude}
In this paper, we thoroughly investigate SVM and DWD on their performance when applied to the HDLSS and imbalanced data.  A novel family of binary classifiers called FLAME is proposed, where SVM and DWD are the two ends of the spectrum. On the DWD end, the estimation of the intercept term is deteriorated while it provides better estimation of the direction vector, and thus better handle the HDLSS data. On the hand, SVM is good at estimating the intercept term but not the direction and is subject to overfitting, and thus is more suitable for imbalanced data but not HDLSS data.

We conduct extensive study of the asymptotic properties of the FLAME family in three different flavors, the `$d$ fixed, $n\rightarrow\infty$' asymptotics (Fisher consistency), the `$d$ and $n_+$ fixed, $n_- \rightarrow\infty$' asymptotics (extremely imbalanced data), and the `$n$ fixed, $d\rightarrow\infty$' asymptotics (the HDLSS asymptotics). These results explain the performance we have seen in the simulations and suggest that with a smart choice of $\theta$, FLAME can properly handle both the HDLSS data and the imbalanced data, by improving the estimations of the direction and the intercept term.

The FLAME family can be immediately extended to multi-class classification, as was done for SVM and DWD such as in \citet{Weston1999Support,crammer2000learnability,lee2004multicategory} or \citet{huang2012multiclass}. Another natural extension is variable selection for FLAME.

The FLAME machines generalize the concepts of support vectors. In SVM, support vectors are referred to vectors that sit on or fall into the two hyperplanes corresponding to $u\leq 1$ (or $u\leq 1/\sqrt{C}$ for the modified version of Hinge loss (\ref{h2loss})). In SVM, only support vectors have impacts on the final solution. DWD is the other extreme case where all the data vectors have some impacts. In the presence of imbalanced sample size, the fact that all the data vectors influence the solution cause the optimization to ignore the minority class. The FLAME with $0<\theta<1$ is somewhere in the middle. For FLAME, part of the data vectors, more than the support vectors, but fewer than all the vectors, have impacts. Smart choice of $\theta$ means that one needs to include as many vectors, and as balanced influential samples, as possible. More vectors usually lead to mitigated overfitting, and balanced sample size of the influential vectors from two classes means that the sensitivity issue of the intercept term can be alleviated.

The authors are aware that it is possible to implement a two step procedure to conduct binary linear classification. In the first step, a good direction is found, probably in the fashion of DWD; in the second step, a fine intercept is chosen by borrowing idea of SVM. However, the theoretical properties of such a procedure are known and will be left as a future research direction.

The choice of $\theta$ usually depends on the nature of the data and the scientific context. If the users prefer better classification performance over reasonable discrimination direction for interpretation of the data, $\theta$ may be chosen to be closer to 1. If the right direction is the first priority, then $\theta$ should be chosen to be closer to 0. Note that, under some circumstances,  the primary goal is to obtain a direction vector which can provide a score $\xv^T\omegav$ for each observation for further use, and the intercept parameter $\beta$ is of no use at all. For example, some users may use a receiver operating characteristic (ROC) curve as a graphical tool to evaluate classification performance over different $\beta$ value instead of using a single $\beta$ value given by the classifier. In this case, a FLAME machine close to the DWD method may be ideal.

\section*{Appendix}\label{sec: appendices}
\setcounter{figure}{0} \setcounter{table}{0} \setcounter{equation}{0}
\makeatletter \renewcommand{\thefigure}{A.\@arabic\c@figure} \renewcommand{\thetable}{A.\@arabic\c@table}   \renewcommand{\theequation}{A.\@arabic\c@equation}
\makeatother

\subsection*{Derivation of the modified Hinge loss}
Note that the original SVM formulation is $\ds{\argmin_{\tilde\omegav,\tilde\beta} \sum \left(1-y_i \tilde f(\xv_i)\right)_+, \mb{s.t. } \|\tilde\omegav\|^2\leq C},$ where $\tilde f(\xv) = \xv^T\tilde\omegav+\tilde\beta$. Here the coefficient vector $\tilde\omegav$ does not have unit norm. We let $\omegav = \tilde\omegav/\sqrt{C}$, $\beta = \tilde\beta/\sqrt{C}$ and $f = \tilde f/\sqrt{C}$. Thus SVM solution is given by
$\ds\argmin_{\omegav,\beta} \sum \left(1-\sqrt{C}y_i f(\xv_i)\right)_+,$ s.t. $\|\omegav\|^2\leq 1,$
or equivalently,
$\ds\argmin_{\omegav,\beta} \sum \left(\sqrt{C}-Cy_i f(\xv_i)\right)_+, \mb{s.t. } \|\omegav\|^2\leq 1.$


\section*{Acknowledgements}
The first author's work was partially supported by Binghamton University Harpur College Dean's New Faculty Start-up Funds and a collaboration grant from the Simons Foundation (\#246649 to Xingye Qiao).

\bibliographystyle{asa}
\bibliography{FLAME_bib}

\clearpage
\newpage
\section*{Online supplementary materials}
\setcounter{figure}{0}
\setcounter{table}{0}
\setcounter{equation}{0}
\makeatletter
\renewcommand{\thefigure}{S.\@arabic\c@figure}
\renewcommand{\thetable}{S.\@arabic\c@table}   
\renewcommand{\theequation}{S.\@arabic\c@equation}
\renewcommand{\thetheorem}{S.\@arabic\c@theorem}
\makeatother
  
\pagenumbering{roman}

This document provides some additional details for the main paper, \textit{Flexible High-dimensional Classification Machines and Their Asymptotic Properties}. This document includes how we implement the FLAME machine with pre-defined $\theta$, detailed proofs for several theorems and propositions, and figures for additional simulations.

\subsection*{Implementation}
In order to implement the FLAME algorithm, we introduce several new notations. Let
$S_{d+1}$ be a second order cone in the $d+1$ dimensional space, $S_{d+1}=\set{(t_0, t_1,
\cdots, t_d)': t_0\geq \sqrt{\sum_{i=1}^d t_i^2}}$. Note that $r_i$ and ${1}/{r_i}$ can be substituted
by three axillary variables $\rho_i$, $\sigma_i$ and $\tau_i$ which satisfy $\rho_i+\sigma_i=r_i$, $\rho_i-\sigma_i=1/r_i$, and $\tau_i=1$. Then $\rho_i^2=\sigma_i^2+\tau_i^2$, and thus $(\rho_i,\sigma_i,\tau_i)\in S_3$. Let $w=1$, then $(w;\omegav)\in S_{d+1}$ since $\|\omegav\|\leq 1$. Let $\eta_i\geq0$,
$\varphi_i\geq0$, where $\varphi_i$ and $\eta_i$ can be viewed as the positive and negative parts of $\Big(\frac{1}{r_i}+C\xi_i-\theta\sqrt{C}\Big)$, \ie, $\varphi_i-\eta_i=\Big(\frac{1}{r_i}+C\xi_i-\theta\sqrt{C}\Big)$.
With the reparameterization above, FLAME can be viewed as the following optimization problem:
\begin{align*}
\min_{\beta,w,\omegav,\rho_i,\sigma_i,\tau_i,\xi_i,\eta_i,\varphi_i}\sum_{i=1}^n \varphi_i\quad\quad\quad\quad& \\
\mb{s.t.}\quad\quad\quad y_i(\xv_i^T\omegav+\beta)+\xi_i-\rho_i-\sigma_i&=0\\
\rho_i-\sigma_i+C\xi_i-\theta\sqrt{C}+\eta_i-\varphi_i&=0\\
w&=1\\
\tau_i&=1\\
\textnormal{and }(w;\omegav)\in S_{d+1}, (\rho_i, \sigma_i, \tau_i)'\in S_3, \xi_i\geq0, \eta_i \geq
0,\varphi_i&\geq 0.
\end{align*}

Therefore, all the constraints can be converted to linear forms, all the variables are either nonnegative, free, or in second order cones, and the objective function is linear. Such problem is called Second Order Cone Programming (SOCP), and can be efficiently solved by softwares such as SDPT3 \citep{Toh1999SDPT,Tutuncu2003Solving}.


\subsection*{Proof to Theorem \ref{thm1:algorithm_converge}}
It suffices to show that $s(\omegav_k,\beta_k,\theta_k)\geq
s(\omegav_{k+1},\beta_{k+1},\theta_{k+1})$. First, $s(\omegav_k,\beta_k,\theta_k)\geq
s(\omegav_k,\beta_k,\theta_{k+1})$ due to the definition of $\theta_{k+1}$ and that
$\theta_k\leq \theta_{k+1}$. Then $s(\omegav_k,\beta_k,\theta_{k+1})\geq
s(\omegav_{k+1},\beta_{k+1},\theta_{k+1})$ since $\omegav_{k+1}$ and $\beta_{k+1}$ minimize
$s(\omegav,\beta,\theta_{k+1})$.\hfill\qed

\subsection*{Proof to Proposition \ref{prop2:fisher}}
For any $\xv$, denote $p(\xv) = \Pr(Y=+1\mid \Xv=\xv)$. The conditional risk is
$R(f)\equiv E[L(Yf(\Xv),\theta)\mid \Xv=\xv]=L(f(\xv),\theta)p(\xv)+L(-f(\xv),
\theta)(1-p(\xv))$. For simplicity, we write
$L(f(\xv), \theta)$ as $L(f)$. Thereby, $R(f)=L(f)p(\xv)+L(-f)(1-p(\xv))$.

We can see that for fixed $p(\xv)\in (0,1)$,  $R(f)$ is continuous and differentiable everywhere
and convex. Thus we find $f^*(\xv)$ by solving $R'(f)=0$, where
$R'(f)=L'(f)p(\xv)+[L(-f)]'(1-p(\xv))$. The FLAME loss is
\begin{align*}
 L(f)=\left\{
	     \begin{array}{cc}
              (2-\theta)\sqrt{C}-Cf & \mb{if } f\leq \frac{1}{\sqrt{C}}\\
	      \frac{1}{f}-\theta\sqrt{C} & \mb{if } \frac{1}{\sqrt{C}}<f\leq
\frac{1}{\theta\sqrt{C}}\\
	      0 & \mb{otherwise}.
             \end{array}\right.
\end{align*}
So direct calculation gives us that
\begin{align*}
 L'(f)=\left\{
	     \begin{array}{cc}
              -C & \mb{if } f\leq \frac{1}{\sqrt{C}}\\
	      -\frac{1}{f^2} & \mb{if } \frac{1}{\sqrt{C}}<f\leq
\frac{1}{\theta\sqrt{C}}\\
	      0 & \mb{otherwise},
             \end{array}\right.
\mb{and } [L(-f)]'=\left\{
	     \begin{array}{cc}
              C & \mb{if } f\geq -\frac{1}{\sqrt{C}}\\
	      \frac{1}{f^2} & \mb{if } -\frac{1}{\theta\sqrt{C}}<f\leq
-\frac{1}{\sqrt{C}}\\
	      0 & \mb{otherwise}.
             \end{array}\right.
\end{align*}
Finally, if $\frac{1-p(\xv)}{p(\xv)}\neq 1$, the solution to $R'(f)=0$ has to be either
$-\frac{1}{\theta\sqrt{C}}<f\leq -\frac{1}{\sqrt{C}}$ or $\frac{1}{\sqrt{C}}<f\leq
\frac{1}{\theta\sqrt{C}}$. In the former case,
$f^*=-\frac{1}{\sqrt{C}}\sqrt{\frac{1-p(\xv)}{p(\xv)}}$ if $\frac{1-p(\xv)}{p(\xv)}>1$. In the
latter case, $f^*=+\frac{1}{\sqrt{C}}\sqrt{\frac{1-p(\xv)}{p(\xv)}}$ if $\frac{1-p(\xv)}{p(\xv)}<1$.
If $\frac{1-p(\xv)}{p(\xv)}=1$, $R'(f^*)=0$ for any $f^*\in
[-\frac{1}{\sqrt{C}},\frac{1}{\sqrt{C}}]$.

Therefore, $f^*$ satisfies
$\sgn(f^*)=\sgn(p(\xv)-0.5)=\sgn(2p(\xv)-1)=\sgn(p(\xv)-(1-p(\xv)))=\sgn(\frac{p(\xv)}{1-p(\xv)}-1).$ \hfill\qed

\subsection*{Proof to Theorem \ref{thm3:oDWD_intercept_diverge}}
Since there are infinitely many negative class samples, it is reasonable to assume that the
classification boundary is pushed closer to the minority positive class, and therefore, the functional margin
$u_i = y_if(\xv_i)=f(\xv_i)$ for the $i$th vector from the minority positive class is small and its DWD loss is $2\sqrt{C}-Cu_i=2\sqrt{C}-Cf(\xv_i)$. Similarly, the DWD loss for the $j$th vector from the majority negative class is $1/[y_jf(\xv_j)]=-1/f(\xv_j)$.  The objective function for DWD is therefore equivalent to $\ds{\frac{1}{n_++n_-}\left\{\sum_{i=1}^{n_+}[2\sqrt{C}-C(\xv_i^T\omegav+\beta)]-\sum_{j=1}^{n_-}\frac{1}{\xv_j^T\omegav+\beta}\right\}+\frac{\lambda}{2}\|\omegav\|^2}.$

The second term inside the curly bracket above can be approximated by\\
$\ds{n_-\int \frac{1}{\xv^T\omegav+\beta}dF_-(\xv)}$ where $F_-(\cdot)$ is the conditional cumulative
distribution function for the negative class. The objective function is therefore
\begin{align*}
l_D=\frac{1}{n_++n_-}\left\{\sum_{i=1}^{n_+}[2\sqrt{C}-C(\xv_i^T\omegav+\beta)]-n_-\int \frac{1}{
\xv^T\omegav+\beta}dF_-(\xv)\right\}+\frac{\lambda}{2}\|\omegav\|^2
\end{align*}

Before we continue, we need the definition that a distribution has a point surrounded \citep{owen2007infinitely}.
\begin{definition}
  The distribution $F$ on $\R^d$ has the point $\xv_*$ surrounded if
  \begin{align*}
  \int_{(\xv-\xv_*)'\omegav>\epsilon}dF(\xv)>\delta,
  \end{align*}
  for some $\delta>0$, some $\epsilon>0$ and all $\omegav\in\R^d$ with $\|\omegav\|=1$.
Consequentially, if $F$ has $\xv_*$ surrounded, then there exist $\gamma$ satisfying
\begin{align}\label{eq:surround2}
 \inf_{\|\omegav\|=1}\int_{(\xv-\xv_*)'\omegav>0}dF(\xv)>\gamma\geq 0.
\end{align}
\end{definition}

We observe that
\begin{align*}
\frac{\partial l_D}{\partial \beta} &=
\frac{1}{n_++n_-}[-n_+C+n_-\int(\xv^T\omegav+\beta)^{-2}dF_-(\xv)]\\
&\geq
\frac{1}{n_++n_-}[-n_+C+n_-\int_{(\xv-\overline\xv_+)'\omegav\geq 0}((\xv-\overline\xv_+)^T\omegav+\overline\xv_+^T\omegav+\beta)^{-2}
dF_-(\xv)]\\
&\geq \frac{1}{n_++n_-}[-n_+C+n_-\int_{\xv^T\omegav\geq 0}(\overline\xv_+^T\omegav+\beta)^{-2}
dF_-(\xv)]\\
&\geq  \frac{1}{n_++n_-}[-n_+C+n_-\frac{\gamma}{(\overline\xv_+^T\omegav+\beta)^2}]
\end{align*}
Now suppose that $-\sqrt{\frac{n_-\gamma}{n_+C}}<\overline\xv_+^T\omegav+\beta<0$, then $\frac{n_-\gamma}{(\overline\xv_+^T\omegav+\beta)^2}>n_+C$ and
$\partial l_D/\partial \beta>0$. Given the fact that $l_D$ is a strictly convex function, the
minimizer $\wh\beta<-\sqrt{\frac{n_-\gamma}{n_+C}}-\overline\xv_+^T\omegav$. \hfill\qed

\subsection*{Proof to Theorem \ref{thm4:w_beta_FLAME}}
Again, with the imbalance assumption we assume that the functional margins for the minority positive class are always greater than 0. Note that the penalized empirical loss for the FLAME machine is approximated by
\begin{align*}
l_F=&\frac{1}{n_++n_-}\left\{\sum_i^{n_+}\left[2\sqrt{C}-C(\xv_i^T\omegav+\beta)-\theta\sqrt{C}\right]
+\right.\\
&\quad \left. n_-\int\left(-
\frac{1}{\xv^T\omegav+\beta}-\theta\sqrt{C}\right)_+dF_-(\xv)\right\}+\frac{\lambda}{2}
\|\omegav\|^2
\end{align*}

Let $g_j^{*}=-(\xv_j^T\omegav^*+\beta^*)$, $j=1, 2, \cdots n_-$ be the functional margins for the negative class. Because $\frac{1}{g_{(n_+)}^{*}\sqrt{C}}=\theta^*$, that is, the reduced loss for the $j$th sample is greater than or equal to 0, $\frac{1}{g_{(n_+)}^{*}}-
\theta^*\sqrt{C}=0$, we observe that $1/g_{(n_+)}^{*}$ is the $n_+$-th greatest among
all the function margins of the negative class $1/g_j^{*}=-1/(\xv_j^T\omegav^*+\beta^*)$. Thus there are at most $n_+$ samples whose reduced losses that are $\geq 0$. Assume that there are $n^o\leq n_+$ such samples.

For a random sample $(\Xv,Y)$ from the negative class, let $E$ be the event that
$(Y(\Xv^T\omegav^*+\beta^*))^{-1}\geq\theta^*\sqrt{C}$. From the argument above,
$\Pr(E)$ is approximately ${n^o}/{n_-}$.

Then the integration $\ds{\int\left(-
\frac{1}{\xv^T\omegav+\beta}-\theta\sqrt{C}\right)_+dF_-(\xv)}$ equals
\begin{align*}
 &\E\left[(-
\frac{1}{\xv^T\omegav+\beta}-\theta\sqrt{C})_+ \mid \overline E\right]\Pr(\overline E)+\E\left[(-
\frac{1}{\xv^T\omegav+\beta}-\theta\sqrt{C})_+\mid E\right]\Pr(E)\\
\approx&\E\left[0\mid\overline E\right](1-\frac{n^o}{n_-})+\E\left[(-
\frac{1}{\xv^T\omegav+\beta}-\theta\sqrt{C})\mid E\right]\frac{n^o}{n_-}\\
=&\E\left[(-\frac{1}{\xv^T\omegav+\beta}-\theta\sqrt{C})\mid E\right]\frac{n^o}{n_-}
\end{align*}

We then have
\begin{align*}
l_F=&\frac{1}{n_++n_-}\left\{\sum_i^{n_+}\left[2\sqrt{C}-C(\xv_i^T\omegav+\beta)-\theta\sqrt{C}\right]
+\right.\\
&\quad\left.n^o\int\left(-\frac{1}{\xv^T\omegav+\beta}-\theta\sqrt{C}\right)dF_-(\xv\mid E)\right\}+\frac{\lambda}{2}\|\omegav\|^2
\end{align*}
Here, $dF_-(\xv\mid E)$ is the conditional distribution function of $\Xv$ for the negative class given event $E$.

Setting $\ds{\partial l_F/\partial\beta=0=\frac{1}{n_++n_-}\left\{-Cn_+
+n^o\int(\xv^T\omegav^*+\beta^*)^{-2}dF_-(\xv\mid E)\right\}},$ we have\\ $\ds{\int(\xv^T\omegav^*+\beta^*)^{-2}dF_-(\xv\mid E)=C\frac{n_+}{n^o}}$.

Setting $\ds{\partial l_F/\partial\omegav=\0v=\frac{1}{n_++n_-}\left\{-Cn_+\overline\xv_+
+n^o\int(\xv^T\omegav^*+\beta^*)^{-2}\xv dF_-(\xv\mid E)\right\}+\lambda\omegav^*},$ we have $\ds{\int(\xv^T\omegav^*+\beta^*)^{-2}\xv dF_-(\xv\mid E)=-\frac{n_++n_-}{n^o}\lambda\omegav^*+C\frac{n_+}{n^o}\overline\xv_+}$. And furthermore,
$$\ds{\frac{\int(\xv^T\omegav^*+\beta^*)^{-2}\xv dF_-(\xv\mid E)}{\int(\xv^T\omegav^*+\beta^*)^{-2}dF_-(\xv\mid E)}=-\frac{n_++n_-}{n_+}\frac{\lambda}{C}\omegav^*+\overline\xv_+=-(1+m)\frac{\lambda}{C}\omegav^*+\overline\xv_+}.$$

That is,
$$\ds{\omegav^*=\frac{C}{(1+m)\lambda}\left[\overline\xv_+-\frac{\int(\xv^T\omegav^*+\beta^*)^{-2}\xv dF_-(\xv\mid E)}{\int(\xv^T\omegav^*+\beta^*)^{-2}dF_-(\xv\mid E)}\right]}
$$\hfill\qed

\subsection*{Proof to Theorem \ref{thm5:w_beta_SVM}}
For simplicity we use the original SVM formulation with the Hinge loss function instead of the FLAME formulation. The objective function for SVM is equivalent to
\begin{align*}
	l_S=&\frac{1}{n_++n_-}\left\{\sum_{i=1}^{n_+}[1-(\xv_i^T\omegav+\beta)]+n_-\int \left[1+(\xv^T\omegav+\beta)\right]_+ dF_-(\xv)\right\}+\frac{\lambda}{2}\|\omegav\|^2\\
	=&\frac{1}{n_++n_-}\left\{\sum_{i=1}^{n_+}[1-(\xv_i^T\omegav+\beta)]+\right.\\
	&\quad\left.n_-\int \left[1+(\xv^T\omegav+\beta)\right]\Ind{1+\xv^T\omegav+\beta>0} dF_-(\xv)\right\}+\frac{\lambda}{2}\|\omegav\|^2
\end{align*}
Setting ${\partial{l_S}}/{\partial{\beta}}=0$, we have
\begin{align*}
\frac{\partial{l_S}}{\partial{\beta}}&=\frac{1}{n_++n_-}\left\{-n_++n_-\Pr(G;\omegav,\beta)+n_-\int(1+\xv^T\omegav+\beta)\delta\left({1+\xv^T\omegav+\beta}\right)dF_-(\xv)\right\}\\
&=\frac{1}{n_++n_-}\left\{-n_++n_-\Pr(G;\omegav,\beta)\right\}=0,
\end{align*}
where $\delta(\cdot)$ is the Dirac delta function.
$$\Rightarrow\Pr(G;\wh\omegav,\wh\beta)=\frac{n_+}{n_-}=\frac{1}{m}.$$

Moreover,
\begin{align*}
\frac{\partial{l_S}}{\partial{\omegav}}=&\frac{1}{n_++n_-}\left\{-\sum_{i=1}^{n_+}\xv_i
+n_-\int\xv\Ind{1+\xv^T\omegav+\beta>0} dF_-(\xv)\right.\\
&+\left. n_-\int \left[1+(\xv^T\omegav+\beta)\right]\delta\left(1+\xv^T\omegav+\beta\right)\xv dF_-(\xv)\right\}+\lambda\omegav\\
=&\frac{1}{n_++n_-}\left\{-n_+\overline\xv_++n_-\int\xv\Ind{1+\xv^T\omegav+\beta>0}dF_-(\xv)\right\}+\lambda\omegav
\end{align*}
\begin{align*}
\Rightarrow \wh\omegav=& \frac{1}{(n_++n_-)\lambda} \left\{n_+\overline\xv_+-n_-\int \xv\Ind{1+\xv^T\wh\omegav+\wh\beta>0}dF_-(\xv)\right\}\\
=& \frac{1}{(n_++n_-)\lambda}\left\{n_+\overline\xv_+-n_-\int \xv  dF_-(\xv|G)\Pr(G;\wh\omegav,\wh\beta)\right\}\\
\approx& \frac{n_+}{(n_++n_-)\lambda}\left\{\overline\xv_+-\int \xv  dF_-(\xv|G)\right\}\\
=& \frac{1}{(1+m)\lambda}\left\{\overline\xv_+-\int \xv dF_-(\xv|G)\right\}.
\end{align*}\hfill\qed

\subsection*{Classification boundaries for HDLSS data}
The geometric representation in \citet{hall2005geometric} leads to some theoretical properties of several binary classifiers. In particular, as $d\rightarrow \infty$, the positive class and negative class converge to two $(n_+-1)$ and $(n_--1)$ simplices with random rotation. Note that the (normalized) pairwise distances between observations within each class are the same, and the (normalized) distances between any two observations from two different classes are the same as well. The geometric representation for SVM and DWD in \citet{hall2005geometric} is summarized as the follows.

\begin{enumerate}
\item \textbf{SVM}: It was shown that the linear SVM hyperplane projected to the $(N-1)$-dimensional subspace that is generated by the $N$ data vectors is given asymptotically by the unique $(N-2)$-dimensional hyperplane that bisects each of the edges of length $l$ in the $N$-polyhedron formed by the $N$ data vectors. There are $n_+\times n_-$ such edges. Let $O^+$ be the centroid of the $(n_+-1)$-simplex $\Xc^+(d)$ and $O^-$ the centroid of the $(n_--1)$-simplex $\Xc^-(d)$. It can be further shown that the SVM hyperplane bisects the line segment between $O^+$ and $O^-$.

\item \textbf{DWD}: The case of DWD is a little different, especially in the case where $n_+\ll n_-$ (or $m\gg 1$), which is our main focus here. We assume that the DWD hyperplane intersects $O^+O^-$ at point $P$. It can be shown that the two simplices, the DWD hyperplane, and the SVM hyperplane, are all orthogonal to $O^+O^-$. Thus all the vertices in the simplex $\Xc^+$ are equally distanced from the DWD hyperplane. Such distance is denoted by $a$. Similarly, all the vertices in the simplex $\Xc^-$ are equally distanced from the DWD hyperplane by $\beta$. The general version DWD hyperplane minimizes the sum of the reciprocals of the distances of data vectors to the hyperplane, $(n_+/a+n_-/b)$, with the constraint that $a+b$ equals to a constant (determined by $\mu,~\sigma,~\tau,~n_+,~n_-,~\mb{and}~d$). A simple calculus practice reveals that $a/b=(n_+/n_-)^{1/2}$.
\end{enumerate}

For the General FLAME case, we need to learn how the hyperplane moves from the point determined by $a/b=(n_+/n_-)^{1/2}$ on $O^+O^-$ to the midpoint of $O^+O^-$ as $\theta$ grows from 0 (DWD) to 1 (SVM). First, we consider the general version of FLAME which seeks to minimize the sum of losses for all data points, $\sum\left(1/u-\theta\sqrt{C}\right)_+$, where the functional margin $u$ is either $a$ or $\beta$ for samples from the positive or the negative classes respectively. When $\theta=0$, the FLAME hyperplane is determined by $a/b=(n_+/n_-)^{1/2}=m^{-1/2}<1$. Thus $b>a$, that is, the hyperplane is closer to the minority class. We renamed them as $a^0$ and $b^0$ where the superscript ``0'' represents the value of $\theta$.

When $\theta>0$ but smaller than $1/(b^0\sqrt{C})$, then the hyperplane does not move. This is because that the new loss for each data vector becomes $1/a^0-\theta\sqrt{C}$ or $1/b^0-\theta\sqrt{C}$ since both are greater than 0. The additional term ``$-\theta\sqrt{C}$'' does not change the minimizer and thus $a^{\theta_1}/b^{\theta_1}=(n_+/n_-)^{1/2}$ does remain unchanged.

If we keep increasing $\theta$ so that it becomes greater than $1/(b^0\sqrt{C})$, then if the hyperplane does not move, then the loss for the majority class becomes 0. In this case, there is space for improvement: the hyperplane would move gradually towards the majority class, because this can make the loss on the minority class smaller while keeping the loss on the majority zero. The FLAME hyperplane is determined by $b=1/(\theta\sqrt{C})$.

Finally, as $\theta$ increases, the distance $a$ increases and the distance $\beta$ decreases, until at a point where $a=b$, and both $1/a-\theta\sqrt{C}=1/b-\theta\sqrt{C}<0$. After this point, further increase of $\theta$ will not change the position of the FLAME hyperplane which will remain at the midpoint of $O^+O^-$.
\begin{figure}[!ht]
  \begin{center}
  	 \includegraphics[height=1\linewidth, width=0.7\linewidth, angle=270]{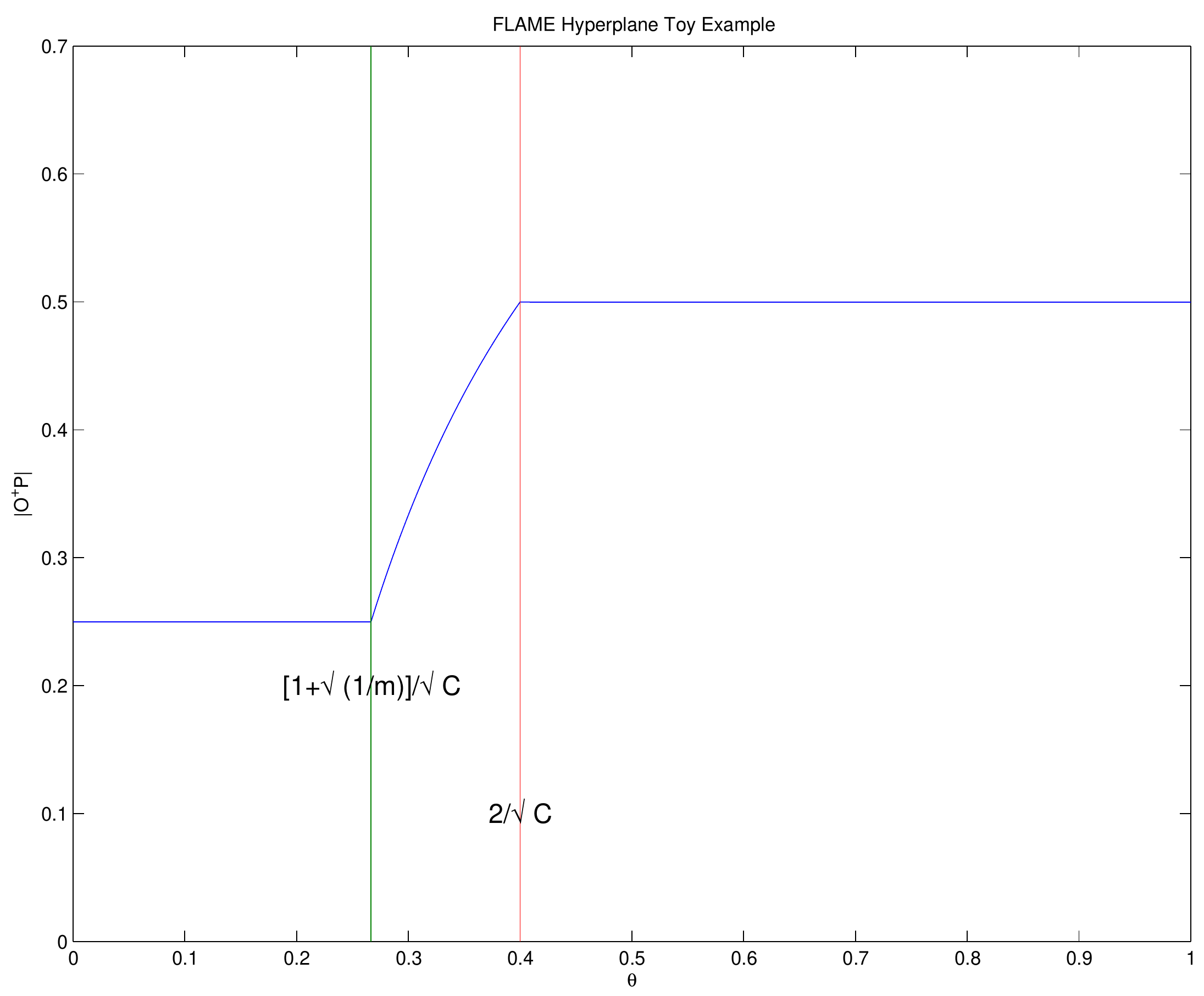}
  \end{center}
  \caption{A 1D toy example with $n_+<n_-$ and $m=9$ is used to mimic the $d$-asymptotic situation. The length of the line segment $O^+O^-$ equals 1. As $\theta$ increases, the FLAME hyperplane stands still ($|O^+P|$ unchanged); when $\theta>\sqrt{1+(1/m)}/\sqrt{C}$,  $|O^+P|$ increases, which means the hyperplane moves towards the negative class, until $\theta=2/\sqrt{C}$, after which the hyperplane remains at the midpoint of $O^+O^-$.}
	\label{fig:hdlss_toy}
\end{figure}

The derivation above assumes the distance between the two simplices are reasonable large, at least greater than $2/\sqrt{C}$. This is not difficult to achieve because we choose $C$ to be a large number.

In summary, the intersection $P$ of the FLAME hyperplane and $O^+O^-$ stays closer to the minority class, and remains still as $\theta$ is small.  When $\theta$ increases, the boundary moves towards the majority class, until reaching the midpoint of $O^+O^-$. This explains the simulation performance we observed in Figures \ref{fig:tryoutIndependent}, \ref{fig:tryoutInterchangeable} and \ref{fig:tryoutBlockCorrelated}. We use a toy example and show the position of the FLAME hyperplane moving as $\theta$ increases in Figure \ref{fig:hdlss_toy} in the same fashion we discussed above.

It is worth noting that the value of DWD/FLAME in terms of reducing overfitting is maximal when the dimension is greater than, but close to, the sample size. This is when data-piling starts to appear in SVM but not yet in DWD. \citet{marron2007distance} showed some videos about such phenomenon. As a matter of fact, according to the geometric representation above, in the $d$-asymptotics, the discriminant directions for most classifiers are the same. Moreover, the projections of data points in the same class to $O^+O^-$ are the same, which is the normal vector for the DWD, SVM and FLAME hyperplanes. Therefore, they all have data-piling in the $d$-asymptotics.

\subsection*{Derivation of the FLAME hyperplane in $d$ asymptotics}
The FLAME seeks to minimize
\begin{align}\label{eq:flame_opt}
&n_+\left(1/a-\theta\sqrt{C}\right)_++n_-\left(1/b-\theta\sqrt{C}\right)_+\\
\mb{s.t.}~&a+b=\sqrt{d(\mu^2+\sigma^2/n_++\tau^2/n_-)}=\nu\sqrt{d}.
\end{align}
When $\theta\in\left[0,(1+\sqrt{m^{-1}})/(\nu\sqrt{dC})\right)$, it is easy to verify that both $(1/a-\theta\sqrt{C})_+$ and $(1/b-\theta\sqrt{C})_+$ are positive and equal to $1/a-\theta\sqrt{C}$ and $1/b-\theta\sqrt{C}$. In this case, the optimal solutions to problem \ref{eq:flame_opt}, $a$ and $\beta$ satisfy $a/b=(n_+/n_-)^{1/2}=\sqrt{m^{-1}}$. In particular, $a=\sqrt{m^{-1}}/(1+\sqrt{m^{-1}})\nu\sqrt{d}$ and $b=1/(1+\sqrt{m^{-1}})\nu\sqrt{d}$.

When $\theta\in\left[(1+\sqrt{m^{-1}})/(\nu\sqrt{dC}),2/(\nu\sqrt{dC}) \right)$, $1/b-\theta\sqrt{C}<0$ and the optimized solutions to (\ref{eq:flame_opt}) are $b=1/(\theta\sqrt{C})$ and $a=\sqrt{d}\nu-b$. Note that $a>b$.

When $\theta\in\left[2/(\nu\sqrt{dC}),1\right]$, $a=b=0.5\sqrt{d}\nu$

\subsection*{Proof to Theorem \ref{thm6:flame_hdlss_cutoff}}
We only need to prove the sure classification for the second interval, \ie, $$\theta\in\left[(1+\sqrt{m^{-1}})/(\nu\sqrt{dC}),2/(\nu\sqrt{dC}) \right).$$ The proofs for the other two intervals are similar to those in \citet{Qiao2010Weighteda}. It was shown in \citet{hall2005geometric} and \citet{Qiao2010Weighteda} that the length of the line segment $O^+O^-$ is $\sqrt{d}\nu$ and that the distance between the projection (denoted as $P'$) of a new data point from the $\Xc^+$-population onto $O^+O^-$ and the centroid of the positive class $O^+$ is $(\sigma^2/n_+)/(\mu^2+\tau^2/n_-)$ times of its distance to the centroid of the negative  class $O^-$, \ie, $\ds{\frac{|O^+P'|}{|O^-P'|}=(\sigma^2/n_+)/(\mu^2+\tau^2/n_-)}$, where $|AB|$ is the length of the line segment connecting points $A$ and $B$. Denote $|O^+P'|$ as $a'$ and $|O^-P'|$ as $b'$. Because $a'+b'=\sqrt{d}\nu$, we must have $b'=\sqrt{d}(\mu^2+\tau^2/n_-)/\nu$. In order for this new data point to be correctly classified to the positive class, $P'$ has to be the same side as $O^+$ with respect to the intersection of the FLAME hyperplane and $O^+O^-$, that is,
\begin{align*}
	&&b'&>b\\
	&\Leftrightarrow& \sqrt{d}\left(\mu^2+\frac{\tau^2}{n_-}\right)/\nu&>\frac{1}{\sqrt{C}\theta}\\
	&\Leftrightarrow& \mu^2+\frac{\tau^2}{n_-}&>\frac{1}{\sqrt{dC}\theta}\nu\\
	&\Leftrightarrow&\nu^2-\frac{\sigma^2}{n_+}&>\frac{1}{\sqrt{dC}\theta}\nu\\
	&\Leftrightarrow&\nu^2-\frac{1}{\sqrt{dC}\theta}\nu-\frac{\sigma^2}{n_+}&>0\\
	 &\Leftrightarrow&(\nu-\frac{1}{2\sqrt{dC}\theta})^2-\frac{1}{4dC\theta^2}-\frac{\sigma^2}{n_+}&>0\\
	 &\Leftarrow&\nu&>\sqrt{\frac{1}{4dC\theta^2}+\frac{\sigma^2}{n_+}}+\frac{1}{2\sqrt{dC}\theta}
	\end{align*}
	\begin{align*}
					 &\Leftrightarrow&\mu^2&>\left[\sqrt{\frac{1}{4dC\theta^2}+\frac{\sigma^2}{n_+}}+\frac{1}{2\sqrt{dC}\theta}\right]^2-\frac{\sigma^2}{n_+}-\frac{\tau^2}{n_-}\\
	&\Leftrightarrow&\mu^2&>T-\frac{\tau^2}{n_-}.
\end{align*}

We now assume that $P'$ is the projection of a new data point from the $\Xc^-$-population. In this situation, it can be shown that $a'/b'=(\mu^2+\sigma^2/n_+)/(\tau^2/n_-)$ and thus $\ds{b'=\sqrt{d}\frac{\tau^2/n_-}{\nu}}$. To correctly classify this new data point, we only need to have $b'<b$. That is,
\begin{align*}
	&&\sqrt{d}\frac{\tau^2/n_-}{\nu}&<b=\frac{1}{\theta\sqrt{C}}\\
	&\Leftrightarrow&\frac{\tau^2}{n_-}&<\frac{1}{\theta \sqrt{dC}}\sqrt{\mu^2+\tau^2/n_-+\sigma^2/n_+}
\end{align*}

We only need to show that $\ds{\frac{\tau^2}{n_-}<\frac{1}{\theta \sqrt{dC}}\sqrt{\tau^2/n_-+\sigma^2/n_+}}$. Let $q^2=\tau^2/n_-+\sigma^2/n_+$. We need to show that
\begin{align*}
	&&q^2-\frac{\sigma^2}{n_+}&<\frac{1}{\theta \sqrt{dC}}q\\
	 &\Leftrightarrow&\left(q-\frac{1}{2\theta\sqrt{dC}}\right)^2-\frac{1}{4\theta^2dC}-\frac{\sigma^2}{n_+}&<0\\
	 &\Leftarrow&q&<\sqrt{\frac{1}{4\theta^2dC}+\frac{\sigma^2}{n_+}}+\frac{1}{2\theta\sqrt{dC}}\\
	 &\Leftrightarrow&\frac{\tau^2}{n_-}&<\left[\sqrt{\frac{1}{4\theta^2dC}+\frac{\sigma^2}{n_+}}+\frac{1}{2\theta\sqrt{dC}}\right]^2-\frac{\sigma^2}{n_+}\\
	&\Leftrightarrow&\frac{\tau^2}{n_-}&<T.
\end{align*}
The last inequality is the condition stipulated in the theorem.\hfill\qed


\newpage
\subsection*{Additional figures}
\begin{figure}[!ht]
  \begin{center}
  		\includegraphics[height=1\linewidth, width=0.7\linewidth, angle=270]{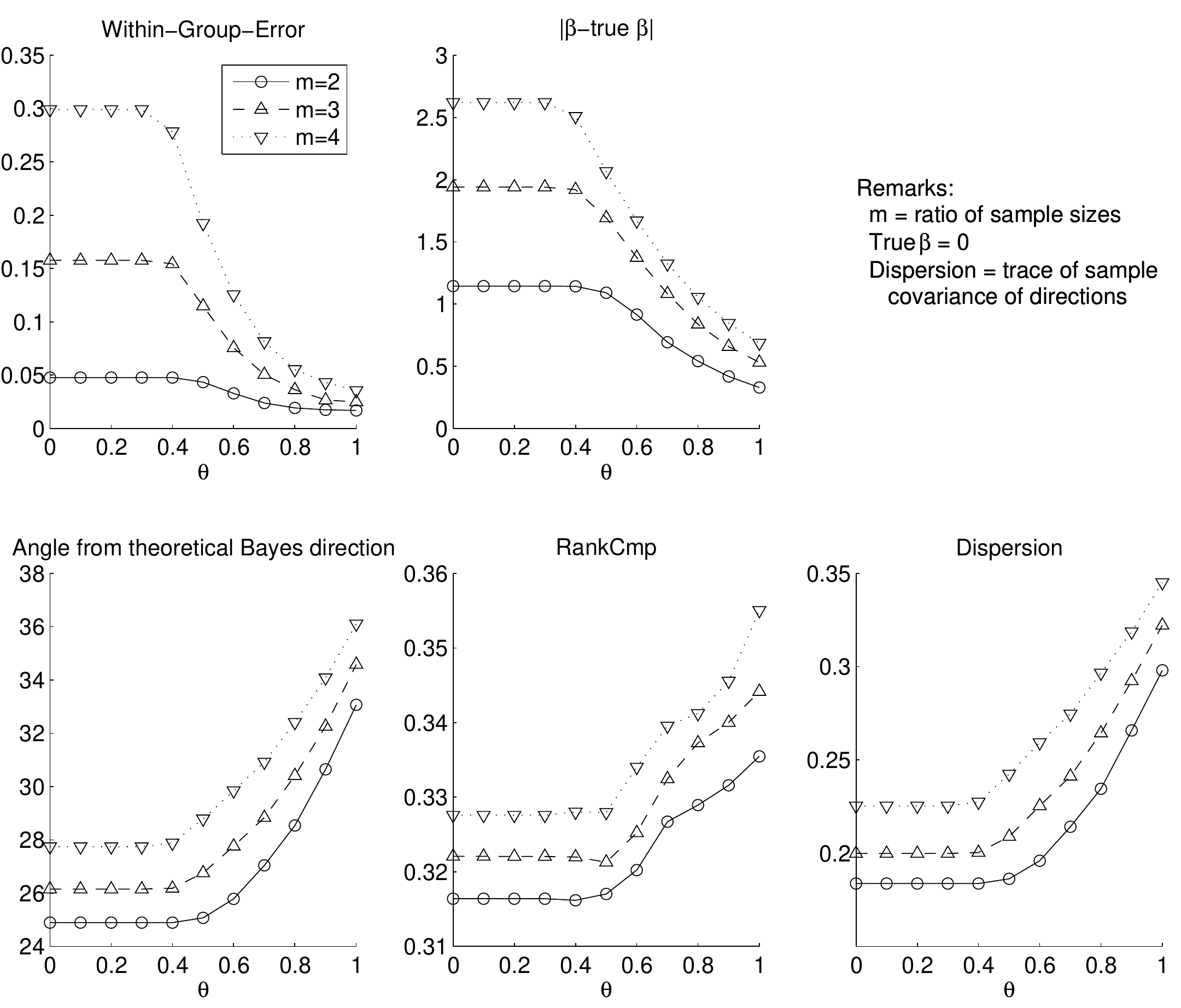}
 	\end{center}
  \caption{Independent example. It can be seen that with FLAME turns from DWD to SVM ($\theta$ from 0 to 1), the within-class error decreases (top-left), thanks to the more accurate estimate of the intercept term (top-middle). On the other hand, this comes at the cost of larger deviation from the Bayes direction (bottom-left), incorrect rank of the importance of the variables (bottom-middle) and larger stochastic variability of the estimation directions (bottom-right).}
  \label{fig:tryoutIndependent}
\end{figure}

\begin{figure}[!p]
  \begin{center}
	   \includegraphics[height=1\linewidth, width=0.7\linewidth, angle=270]{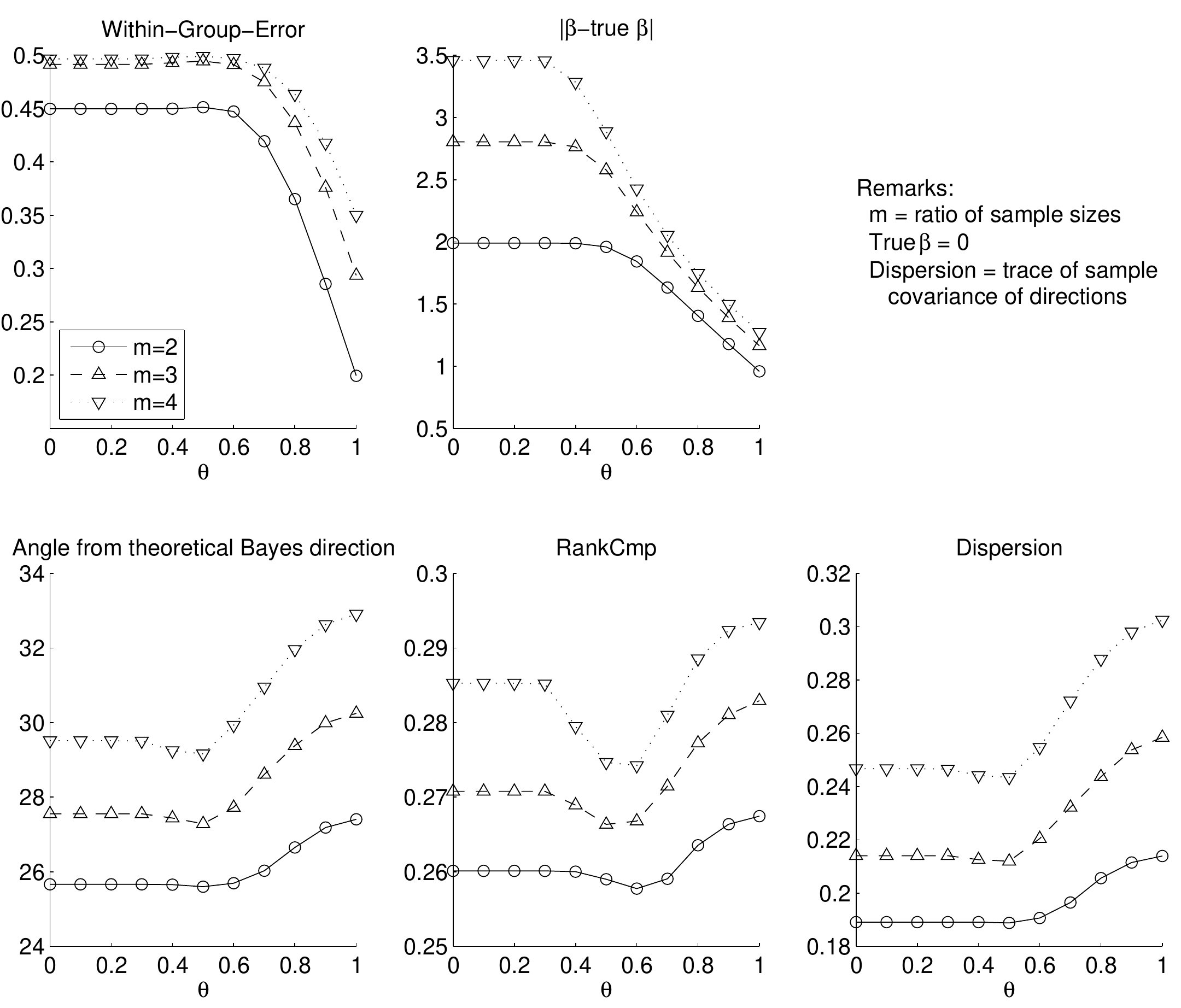}
  \end{center}
  \caption{Block interchangeable example. It can be seen that with FLAME turns from DWD to SVM ($\theta$ from 0 to 1), the within-class error decreases (top-left), thanks to the more accurate estimate of the intercept term (top-middle). On the other hand, this comes at the cost of larger deviation from the Bayes direction (bottom-left), incorrect rank of the importance of the variables (bottom-middle) and larger stochastic variability of the estimation directions (bottom-right). The RankComp measure in this example is an exception, in the sense that it decreases first then increases instead of monotonically increases.}
  \label{fig:tryoutBlockCorrelated}
\end{figure}

\end{document}